\documentclass[conference]{IEEEtran}
\usepackage{times}

\usepackage{times}

\usepackage[numbers]{natbib}
\usepackage{multicol}
\usepackage[bookmarks=true]{hyperref}
\usepackage{xspace}
\usepackage{amsmath}
\usepackage{amssymb}

\usepackage{siunitx}
\usepackage{graphicx}
\usepackage{subcaption}
\usepackage[numbers]{natbib}
\usepackage{booktabs}       
\usepackage{comment}

\usepackage{algorithm} 
\usepackage{algorithmic}  
\usepackage[algo2e]{algorithm2e} 
\usepackage{bm}

\usepackage{xcolor}

\let\labelindent\relax
\usepackage{enumitem}

\usepackage{algorithm2e}

\newcommand{\xxnote}[3]{}
\ifx\hidenotes\undefined
  \usepackage{color}
  \renewcommand{\xxnote}[3]{\color{#2}{#1: #3}}
\fi

\usepackage[normalem]{ulem}

\begin{document}

\title{On the Importance of Environments in Human-Robot Coordination}



\author{
\authorblockN{
Matthew C. Fontaine$^*$,
Ya-Chuan Hsu$^*$, 
Yulun Zhang$^*$, 
Bryon Tjanaka, 
Stefanos Nikolaidis}
\authorblockA{Department of Computer Science\\
University of Southern California\\
Los Angeles, CA, USA\\
mfontain@usc.edu, yachuanh@usc.edu, yulunzha@usc.edu, tjanaka@usc.edu, nikolaid@usc.edu}
}


%

\maketitle
\begingroup\renewcommand\thefootnote{$^*$}
\footnotetext{Equal contribution}
\endgroup

\begin{abstract}
When studying robots collaborating with humans, much of the focus has been on robot policies that coordinate fluently with human teammates in collaborative tasks. However, less emphasis has been placed on the effect of the environment on coordination behaviors. To thoroughly explore environments that result in diverse behaviors, we propose a framework for procedural generation of environments that are (1) stylistically similar to human-authored environments, (2) guaranteed to be solvable by the human-robot team, and (3) diverse with respect to coordination measures.  We analyze the procedurally generated environments in the Overcooked benchmark domain via simulation and an online user study. Results show that the environments result in qualitatively different emerging behaviors and statistically significant differences in collaborative fluency metrics, even when the robot runs the same planning algorithm.
\end{abstract}

\IEEEpeerreviewmaketitle

\section{Introduction}
When humans and robots coordinate well, they time their actions precisely and efficiently and alter their plans dynamically, often in the absence of verbal communication. Evaluation of the quality of coordination has focused not only on task efficiency but on the \textit{fluency} of the interaction~\cite{hoffman2019evaluating}. Fluency refers to how well the actions of the agents are synchronized, resulting in coordinated meshing of joint activities. 

A closely related, important aspect of human-robot teaming is workload assignment. Human factors research has shown that too light or too heavy workload can affect human performance and situational awareness~\cite{parasuraman2008situation}. The perceived robot's contribution to the team is a crucial metric of fluency~\cite{hoffman2019evaluating}, and human-robot teaming experiments found that the degree to which participants were occupied affected their subjective assessment of the robot as a teammate~\cite{gombolay2017computational}.

\begin{figure}[!t]
\centering
\includegraphics[width=\linewidth]{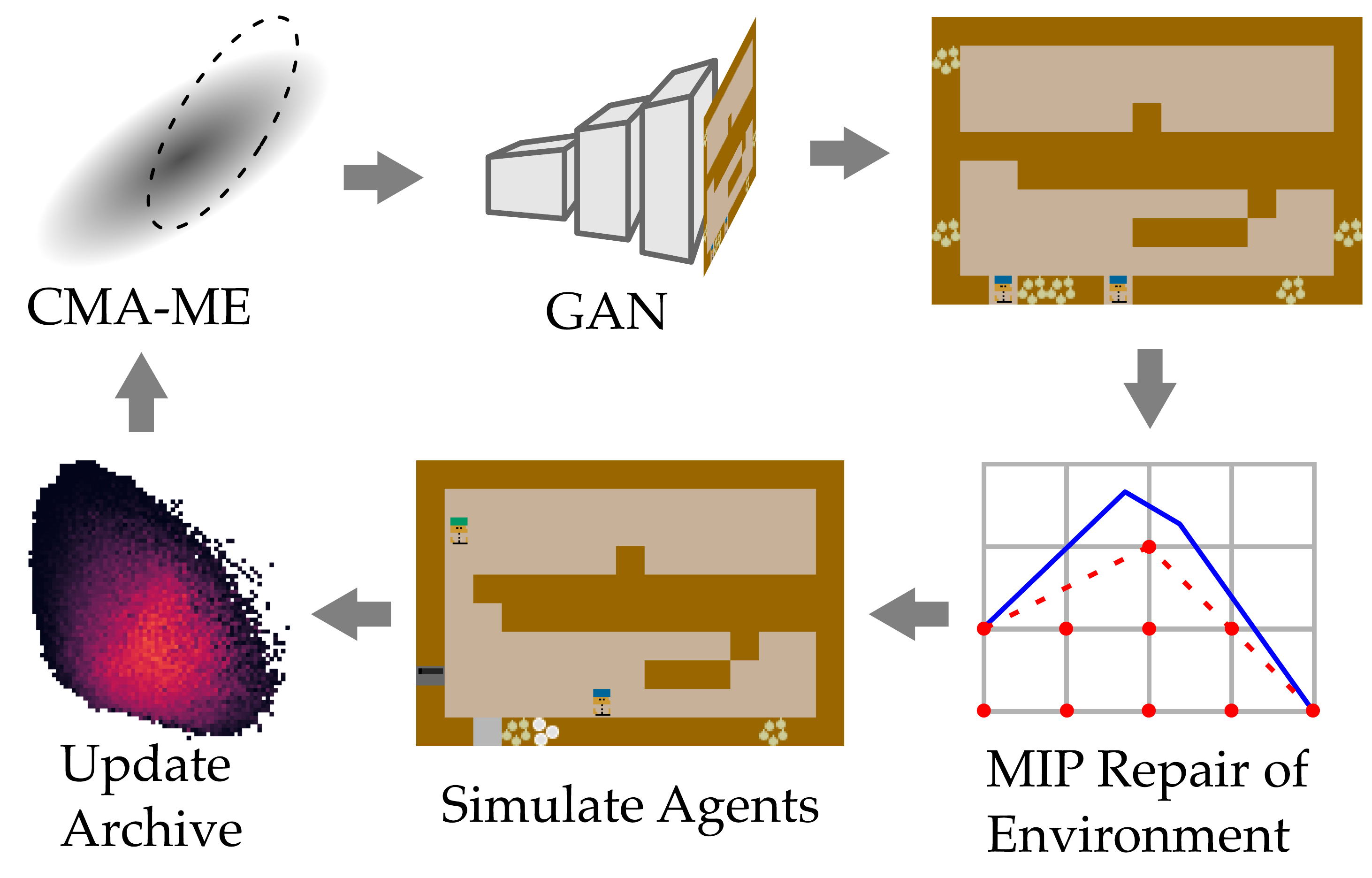}
\caption{An overview of the framework for procedurally generating environments that are stylistically similar to human-authored environments. Our environment generation pipeline enables the efficient exploration of the space of possible environments to procedurally discover environments that differ based on provided metric functions. }
\label{fig:best}
\end{figure}


To achieve fluent human-robot coordination, previous work~\cite{hoffman2019evaluating} enabled robots to reason over the mental states and actions of their human partners, by building or learning human models and integrating these models into decision making. While the focus has been on the effect of these models on human-robot coordination, little emphasis has been placed on the \textit{environment} that the team occupies.  


Our thesis is that \textit{changing the environment can result in significant differences between coordination behaviors}, even when the robot runs the same coordination algorithm. We thus advocate for considering diverse environments when studying the emergent coordination behaviors of human-robot teams.





Manually creating environments that show a diverse range of coordination behaviors requires substantial human effort. Furthermore, as robotic systems become more complex, it becomes hard to predict how these systems will act in different situations and even harder to design environments that elicit a diverse range of behaviors.

This highlights the need for a systematic approach for generating environments. Thus, we propose a framework for automatic environment generation, drawing upon insights from the field of procedural content generation in games~\cite{togelius2014procedural,fontaine2020illuminating,zhang2020video}.
The framework automatically generates environments which (1) share design similarities to human-authored environments provided as training data, (2) are guaranteed to be solvable by the human-robot team, and (3) result in coordination behaviors that differ according to provided metrics (i.e. team fluency or workload).

In this paper, we study our framework in the collaborative game Overcooked~\cite{carroll2019utility}, an increasingly popular domain for researching the coordination of agent behaviors. In this domain, a human-robot team must work together to cook and serve food orders in a shared kitchen. In the context of Overcooked, our framework generates kitchen environments that cause the human-robot team to behave differently with respect to workload, fluency, or any other specified metric. 

Fig.~\ref{fig:best} provides an overview of our framework. First, we train a generative adversarial network (GAN)~\citep{goodfellow:nips14} with human-authored kitchens as examples. The GAN learns to generate kitchens that share stylistic similarities to the human-designed kitchens. However, GANs frequently generate kitchens which are, for the human-robot team, impossible to solve. For example, ingredients or the serve station may be unreachable by the agents, or the kitchen may not have a pot. 
To guarantee the kitchen is solvable, mixed-integer linear programming (MIP)~\citep{zhang2020video} edits the kitchen with a minimum-cost repair. By guaranteeing domain-specific constraints, the \mbox{GAN+MIP} pipeline forms a generative space of viable kitchens that can be sampled through the latent codes of the GAN. We then search the latent space directly with a state-of-the-art \textit{quality diversity} algorithm, Covariance Matrix Adaptation \mbox{MAP-Elites} (\mbox{CMA-ME})~\citep{fontaine2019covariance}, to discover kitchens that cause diverse agent behaviors with respect to specified coordination metrics. 
Generated kitchens are added to an archive organized by the coordination metrics, and  feedback from how the archive is populated helps guide \mbox{CMA-ME} towards kitchens with underexplored metric combinations.

Evaluation of our framework in simulation shows that the generated environments can affect the coordination of human-robot teams that follow precomputed jointly optimal motion plans, as well as of teams where the robot reasons over partially observable human subtasks. In an online user study, we show that our generated environments result in very different workload distributions and team fluency metrics, even when the robot runs the same algorithm in all environments. 




Overall, we are excited to highlight the role that environments play in the emergent coordination behaviors of human-robot teams, and to provide a systematic approach for procedurally generating high-quality environments that induce diverse coordination behaviors.

\section{Background}

\noindent\textbf{Human-Aware Planning.} 
In the absence of pre-coordination strategies, robots coordinate with human teammates by reasoning over human actions when making decisions. 
While many works have studied human-aware planning~\cite{alami2006toward,chakraborti2018human,kruse2013human,mainprice2011planning,broad2018operation}, most relevant to this work are POMDP frameworks, where the robot observes human actions to infer the internal state of a human. POMDP-based models have enabled communication with unknown teammates~\cite{barrett2014communicating}, inference of human preference~\cite{nikolaidis2015efficient} and human trust~\cite{chen2020trust} in human-robot collaboration, and inference of human internal state~\cite{sadigh2016information} in autonomous driving applications.


Since the exact computation of a POMDP policy is computationally intractable~\cite{papadimitriou1987complexity}, researchers have proposed several approximation methods. One such approximation is the QMDP,  where the robot estimates its current actions based on the current belief and the assumption of full observability at the next time step~\cite{littman1995learning}. Though the robot does not take information-gathering actions in this approximation, QMDP has been shown to achieve good performance in domains when the user continuously provides inputs to the system, such as in shared autonomy~\cite{javdani2015shared}. 

Most relevant to our implementation is the CAPIR framework~\cite{nguyen2011capir}, which implements a QMDP agent by decomposing a ghost-hunting game into MDPs that model different sub-tasks (ghosts) that the human may wish to complete. The CAPIR framework was also tested in a Cops and Robbers game~\cite{macindoe2012pomcop}. Given the large number of subtasks in the Overcooked environment, we use a QMDP planner to pick the next subtask, e.g, ``pick an onion'', while the cost of transitioning from each subtask to the next is derived from pre-computed jointly optimal motion plans. 




\noindent\textbf{Procedural Content Generation.} Procedural content generation (PCG) refers to algorithmic, as opposed to manual, generation of  content~\cite{togelius2014procedural}. A growing research area is PCG via machine learning (PCGML)~\cite{summerville2018procedural}, where content is generated with models trained on existing content (e.g.,~\cite{summerville2016super,snodgrass2014experiments,guzdial2016game}). 


Our work builds on the recent success of GANs as procedural generators of video game levels~\citep{volz2018evolving,giacomello:gem18}. However, generating video game levels is not as simple as training a GAN and then sampling the generator since many generated levels are unplayable. Previous work~\citep{volz2018evolving} addressed this by optimizing directly in latent space via Latent Variable Evolution (LVE)~\citep{bontrager:icbtas18}. Specifically, the authors optimize with the Covariance Matrix Adaptation Evolution Strategy (\mbox{CMA-ES})~\citep{hansen:cma16} to find latent codes of levels that maximize playability and contain specific characteristics (e.g., an exact number of player jumps) in the game \textit{Super Mario Bros.}. 


However, game designers rarely know \emph{exactly} which properties they want the generated levels to have. Later work proposed \textit{Latent Space Illumination}~\citep{fontaine2020illuminating, schrum:cppn:geeco20} by framing the search of latent space as a \textit{quality diversity} (QD)~\cite{chatzilygeroudis2020quality} problem instead of a single-objective optimization problem. In addition to an objective function, the QD formulation permits several measure functions which form the behavior characteristics (BCs) of the problem. The authors generated levels that maximized playability but varied in measurable properties, i.e., the number of enemies or player jumps. Their work showed \mbox{CMA-ME}~\citep{fontaine2019covariance} outperformed other QD algorithms~\citep{mouret2015illuminating, vassiliades:gecco18} when directly illuminating latent space. In the case where the objective and measure functions are differentiable, recent work shows that differentiable quality diversity (DQD)~\cite{fontaine2021differentiable} can significantly improve search efficiency.


As stated above, GAN-generated environments, including those generated with LSI, are frequently invalid. For instance, in Overcooked, they may have more than one robot or it may be impossible for players to reach the stove. Previous work~\citep{zhang2020video} proposed a mixed-integer linear programming (MIP) repair method for the game \textit{The Legend of Zelda}, which edits the input levels to satisfy formal playability constraints. The authors formulate the repair as a minimum edit distance problem, and the MIP repairs the level with the minimum total cost of edits possible.


Our work integrates the \mbox{GAN+MIP} repair method with Latent Space Illumination. This allows us to move the objective of LSI away from simply generating valid environments to generating environments that maximize or minimize team performance. Additionally, while previous work on LSI~\cite{schrum:cppn:geeco20,fontaine2020illuminating} generated levels that were diverse with respect to level mechanics and tile distributions, e.g., number of objects, we focus on the problem of diversity in the agent behaviors that emerge from the generated environments.




\begin{figure}[!t]
\centering
\includegraphics[width=0.66\linewidth]{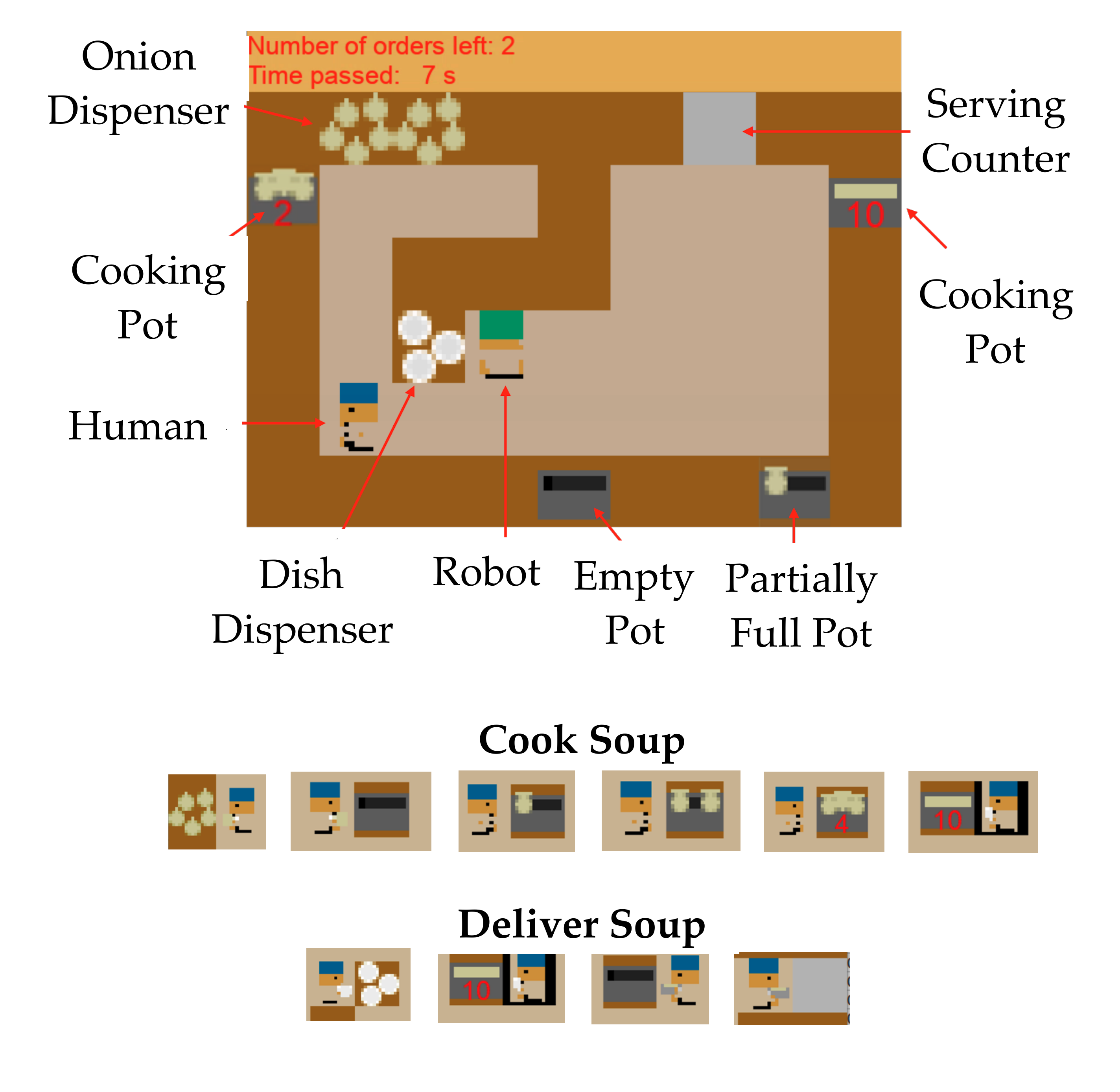}
\caption{Overcooked environment, with instructions for how to cook and deliver a soup.}
\label{fig:instructions}
\end{figure}

\noindent\textbf{Overcooked.} Our coordination domain is based on the collaborative video game \textit{Overcooked}~\citep{overcooked, overcooked2}. Several works~\citep{carroll2019utility, wang2020too} created custom Overcooked simulators as a domain for evaluating collaborative agents against human models~\citep{carroll2019utility} and evaluating decentralized multi-agent coordination algorithms~\citep{wang2020too}. 


 We use the \textit{Overcooked AI} simulator~\citep{carroll2019utility}, which restricts the game to two agents and a simplified item set. In this version, the goal is to deliver two soup orders in a minimum amount of time. To prepare a soup, an agent needs to pick up three onions (one at a time) from an onion dispenser and place them in the cooking pot. Once the pot is full and \num{10} timesteps have passed, the soup is ready. One of the agents then needs to retrieve a dish from the dish dispenser, put the soup on the dish, and deliver it to the serving counter (Fig.~\ref{fig:instructions}).

\begin{figure*}[t!]
\centering

\begin{tabular}{lccccc}
\begin{subfigure}[l]{.08\linewidth}
\begin{flushleft}

\hspace{0.5em}\raisebox{\dimexpr 2.5cm-\height}{  
    Human
    }        \end{flushleft}

    \end{subfigure}
&
\begin{subfigure}[b]{.17\linewidth}
\centering
  \begin{tabular}{cc}
 \resizebox{1.0\linewidth}{!}{
  \includegraphics[width=\linewidth]{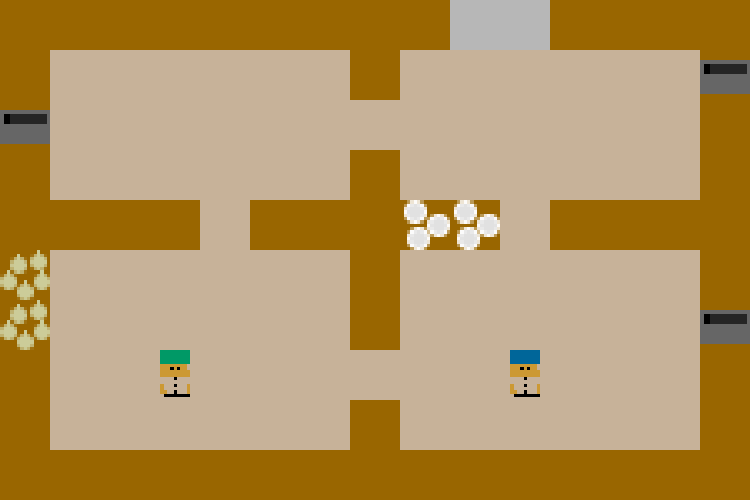}
   }
   \end{tabular}
 \label{fig:plotT0}
\end{subfigure}
\begin{subfigure}[b]{.17\linewidth}
\centering
  \begin{tabular}{cc}
  \resizebox{1.0\linewidth}{!}{
  \includegraphics[width=\linewidth]{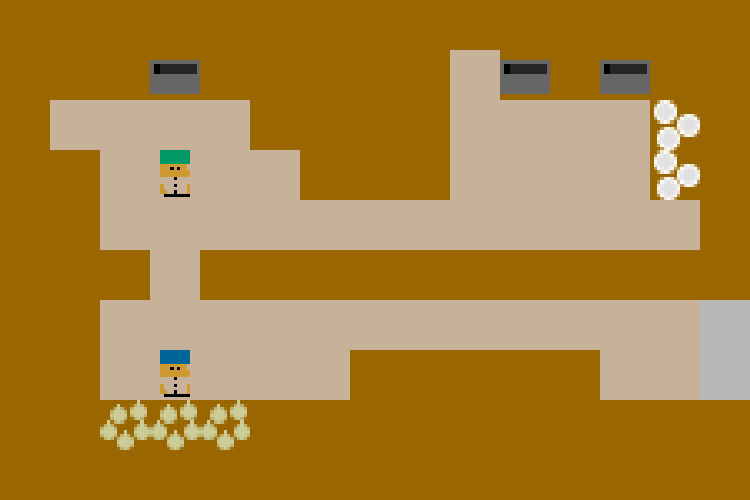}
   }
   \end{tabular}
 \label{fig:plotT0}
\end{subfigure}
\begin{subfigure}[b]{.17\linewidth}
\centering
  \begin{tabular}{cc}
    \resizebox{1.0\linewidth}{!}{
  \includegraphics[width=\linewidth]{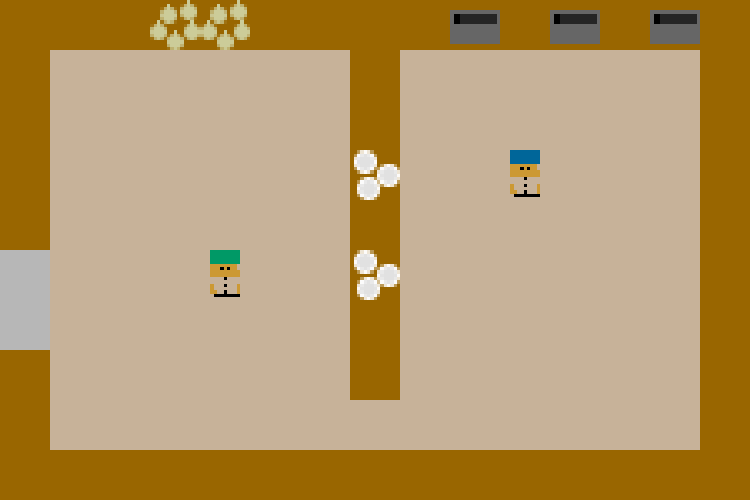}
   }  
   \end{tabular}
 \label{fig:plotT0}
\end{subfigure}
\begin{subfigure}[b]{.17\linewidth}
\centering
  \begin{tabular}{cc}
    \resizebox{1.0\linewidth}{!}{
  \includegraphics[width=\linewidth]{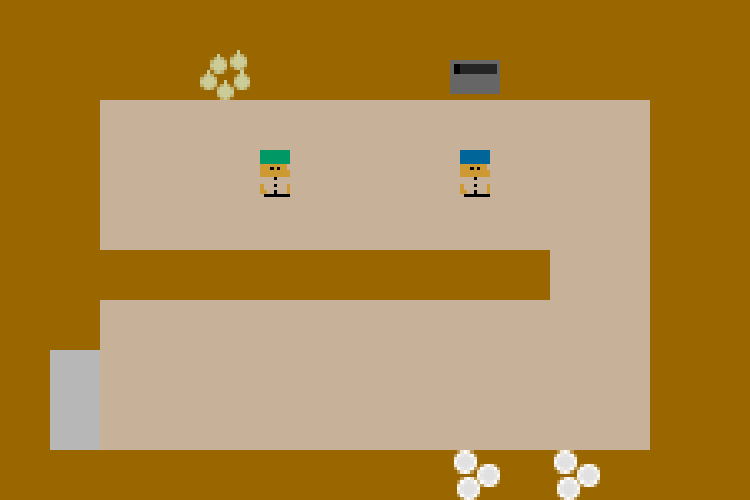}
   }
   \end{tabular}
 \label{fig:plotT0}
\end{subfigure}
\begin{subfigure}[b]{.17\linewidth}
\centering
  \begin{tabular}{cc}
    \resizebox{1.0\linewidth}{!}{
  \includegraphics[width=\linewidth]{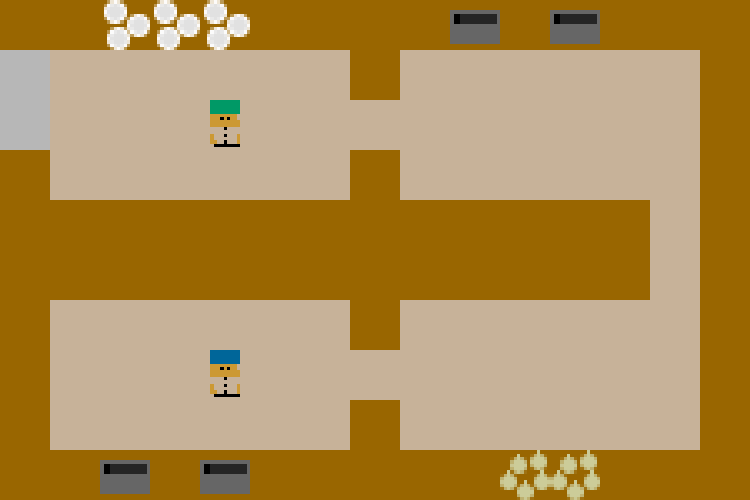}
   }   \end{tabular}
 \label{fig:plotT0}
\end{subfigure}
    \vspace{-1cm}

\end{tabular}

\begin{tabular}{lccccc}
\begin{subfigure}[l]{.08\linewidth}
\centering
    \hspace{0.5em}\raisebox{\dimexpr 2.5cm-\height}{ GAN~~~~~}\end{subfigure}
&
\begin{subfigure}[b]{.17\linewidth}
\centering
  \begin{tabular}{cc}
 \resizebox{1.0\linewidth}{!}{
  \includegraphics[width=\linewidth]{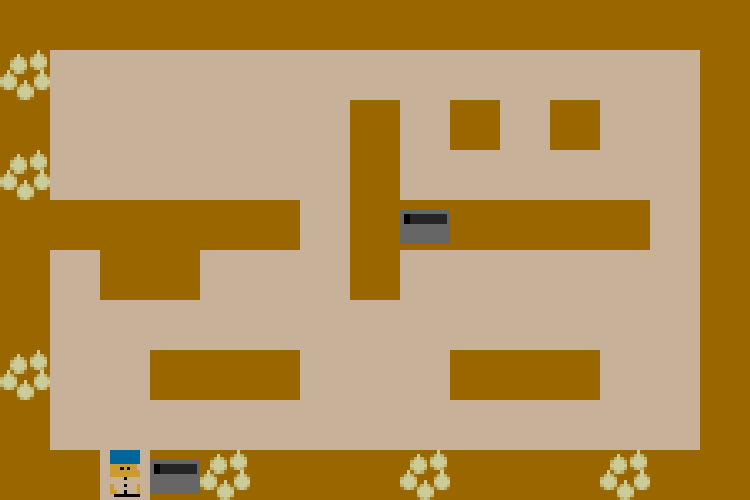}
   }
   \end{tabular}
 \label{fig:plotT0}
\end{subfigure}
\begin{subfigure}[b]{.17\linewidth}
\centering
  \begin{tabular}{cc}
  \resizebox{1.0\linewidth}{!}{
  \includegraphics[width=\linewidth]{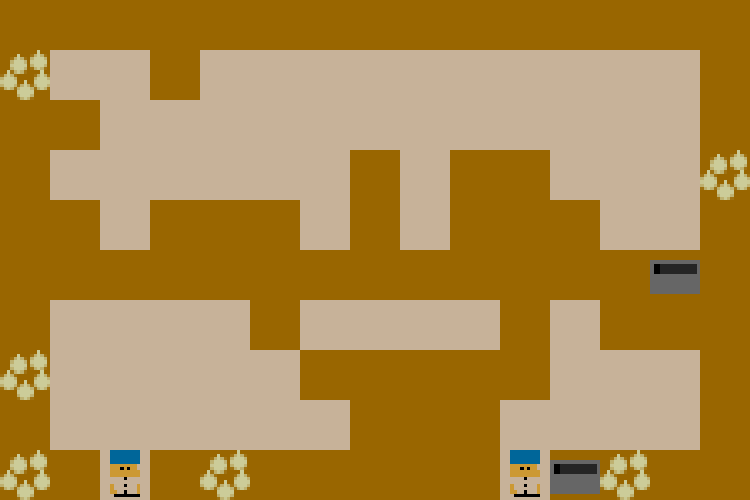}
   }
   \end{tabular}
 \label{fig:plotT0}
\end{subfigure}
\begin{subfigure}[b]{.17\linewidth}
\centering
  \begin{tabular}{cc}
    \resizebox{1.0\linewidth}{!}{
  \includegraphics[width=\linewidth]{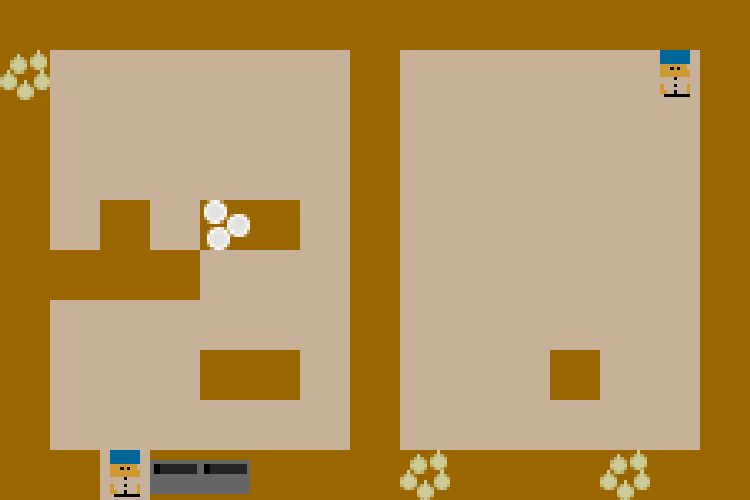}
   }  
   \end{tabular}
 \label{fig:plotT0}
\end{subfigure}
\begin{subfigure}[b]{.17\linewidth}
\centering
  \begin{tabular}{cc}
    \resizebox{1.0\linewidth}{!}{
  \includegraphics[width=\linewidth]{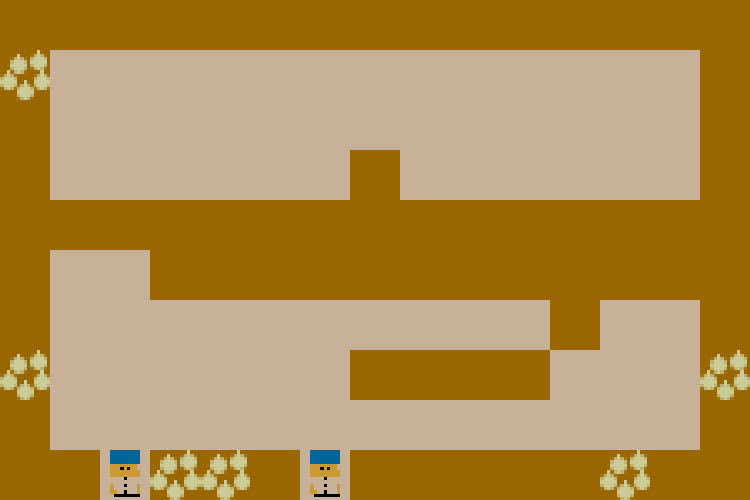}
   }
   \end{tabular}
 \label{fig:plotT0}
\end{subfigure}
\begin{subfigure}[b]{.17\linewidth}
\centering
  \begin{tabular}{cc}
    \resizebox{1.0\linewidth}{!}{
  \includegraphics[width=\linewidth]{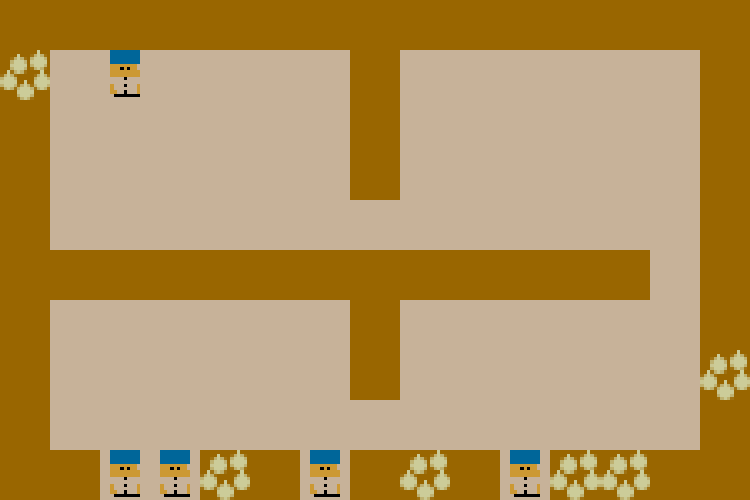}
   }   \end{tabular}
 \label{fig:plotT0}
\end{subfigure}
\vspace{-1cm}

\end{tabular}
\begin{tabular}{lccccc}
\begin{subfigure}[l]{.08\linewidth}
\begin{flushleft}    \hspace{0.5em}\raisebox{\dimexpr 2.5cm-\height}{  
    GAN+MIP~
    }        \end{flushleft}

    \end{subfigure}
&
\begin{subfigure}[b]{.17\linewidth}
\centering
  \begin{tabular}{cc}
 \resizebox{1.0\linewidth}{!}{
  \includegraphics[width=\linewidth]{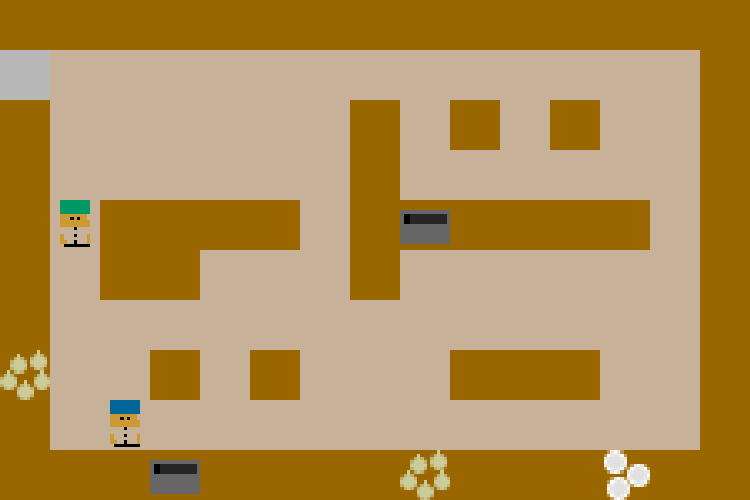}
   }
   \end{tabular}
 \label{fig:plotT0}
\end{subfigure}
\begin{subfigure}[b]{.17\linewidth}
\centering
  \begin{tabular}{cc}
  \resizebox{1.0\linewidth}{!}{
  \includegraphics[width=\linewidth]{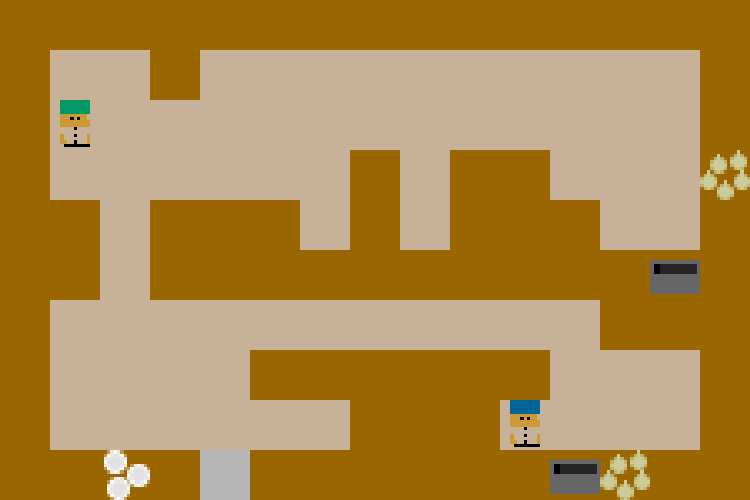}
   }
   \end{tabular}
 \label{fig:plotT0}
\end{subfigure}
\begin{subfigure}[b]{.17\linewidth}
\centering
  \begin{tabular}{cc}
    \resizebox{1.0\linewidth}{!}{
  \includegraphics[width=\linewidth]{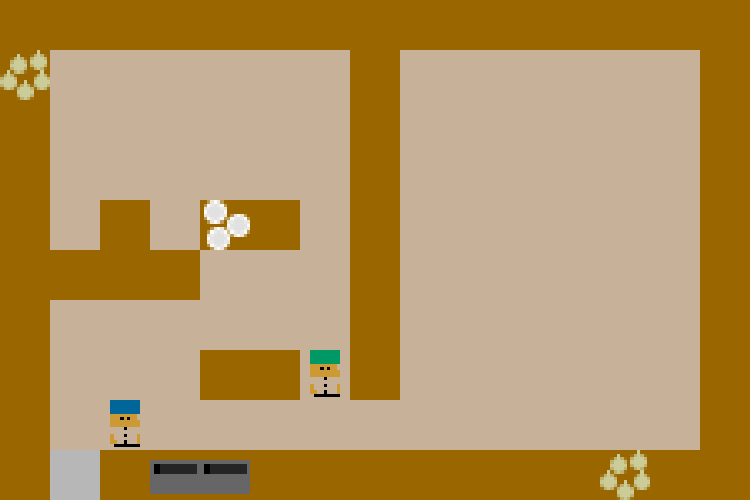}
   }  
   \end{tabular}
 \label{fig:plotT0}
\end{subfigure}
\begin{subfigure}[b]{.17\linewidth}
\centering
  \begin{tabular}{cc}
    \resizebox{1.0\linewidth}{!}{
  \includegraphics[width=\linewidth]{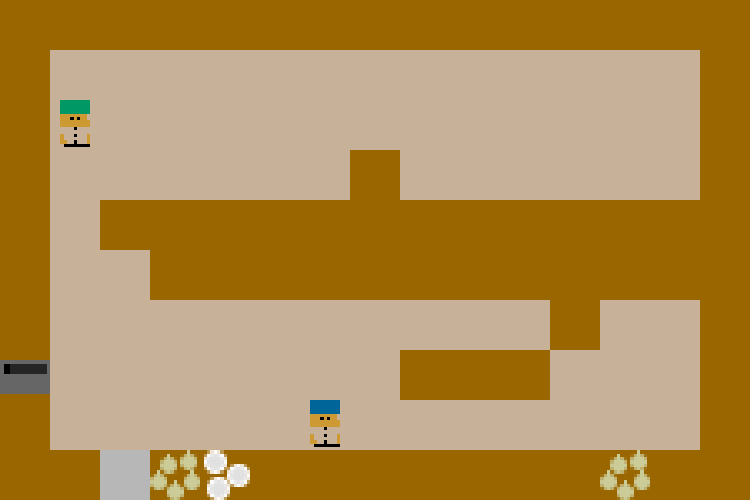}
   }
   \end{tabular}
 \label{fig:plotT0}
\end{subfigure}
\begin{subfigure}[b]{.17\linewidth}
\centering
  \begin{tabular}{cc}
    \resizebox{1.0\linewidth}{!}{
  \includegraphics[width=\linewidth]{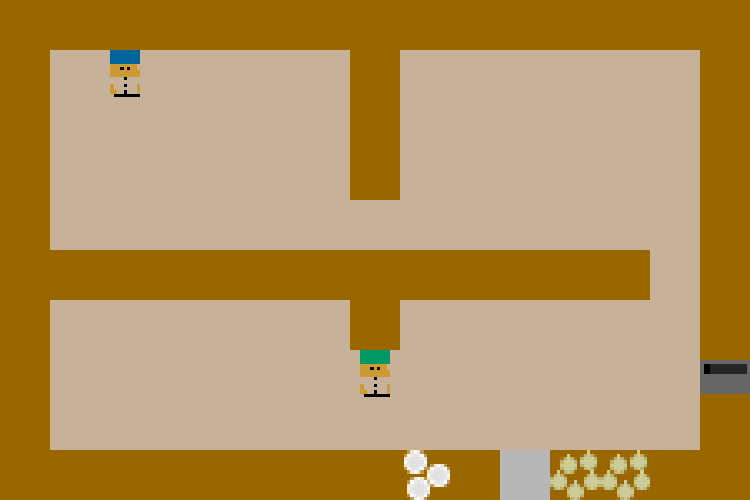}
   }   \end{tabular}
 \label{fig:plotT0}
\end{subfigure}
\vspace{-1cm}
\end{tabular}

\begin{tabular}{lccccc}
\begin{subfigure}[l]{.08\linewidth}
\begin{flushleft}    \hspace{0.5em}\raisebox{\dimexpr 2.5cm-\height}{  
    MIP-random
    }        \end{flushleft}
    \end{subfigure}
&
\begin{subfigure}[b]{.17\linewidth}
\centering
  \begin{tabular}{cc}
 \resizebox{1.0\linewidth}{!}{
  \includegraphics[width=\linewidth]{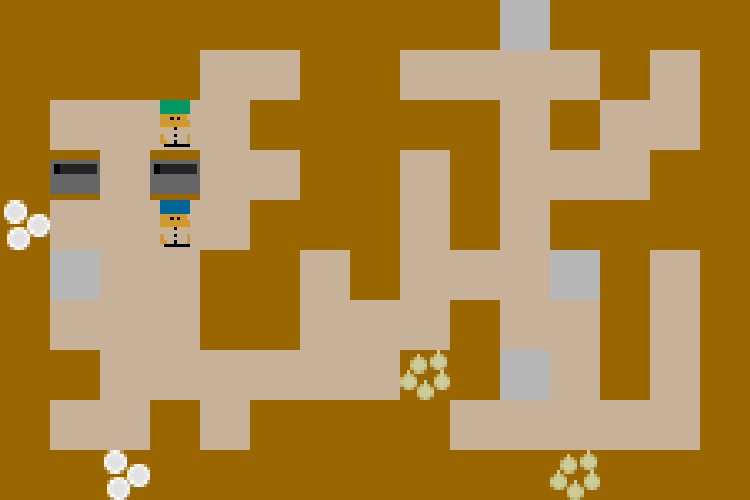}
   }
   \end{tabular}
 \label{fig:plotT0}
\end{subfigure}
\begin{subfigure}[b]{.17\linewidth}
\centering
  \begin{tabular}{cc}
  \resizebox{1.0\linewidth}{!}{
  \includegraphics[width=\linewidth]{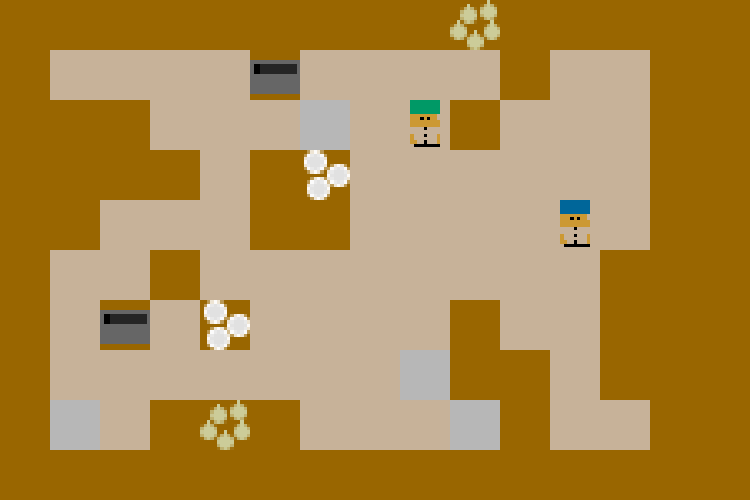}
   }
   \end{tabular}
 \label{fig:plotT0}
\end{subfigure}
\begin{subfigure}[b]{.17\linewidth}
\centering
  \begin{tabular}{cc}
    \resizebox{1.0\linewidth}{!}{
  \includegraphics[width=\linewidth]{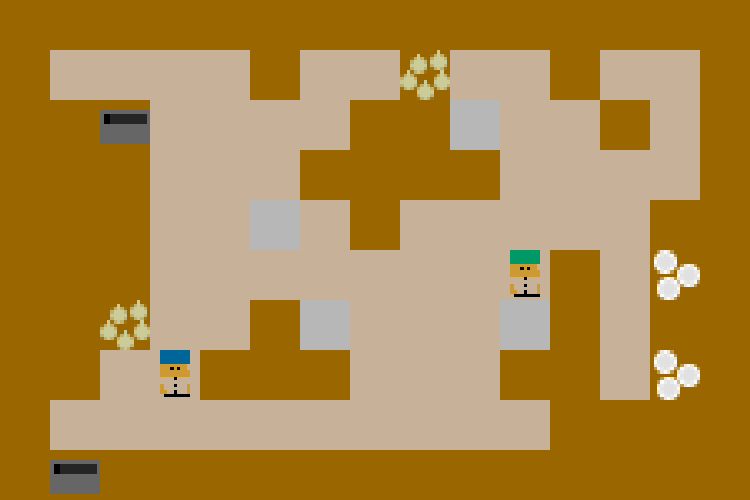}
   }  
   \end{tabular}
 \label{fig:plotT0}
\end{subfigure}
\begin{subfigure}[b]{.17\linewidth}
\centering
  \begin{tabular}{cc}
    \resizebox{1.0\linewidth}{!}{
  \includegraphics[width=\linewidth]{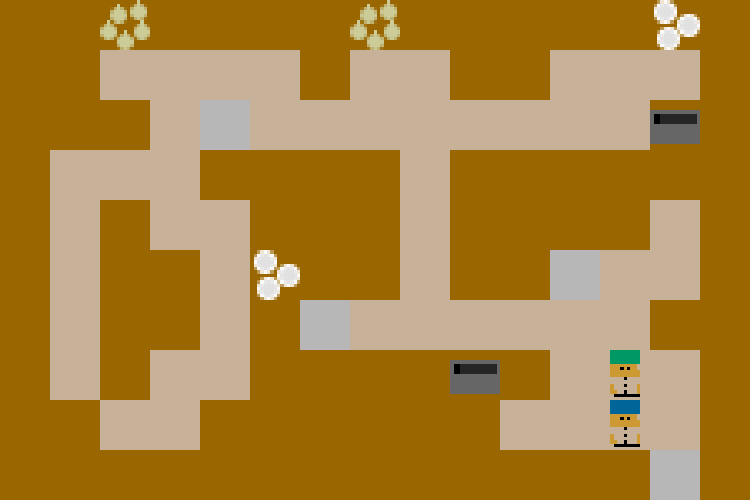}
   }
   \end{tabular}
 \label{fig:plotT0}
\end{subfigure}
\begin{subfigure}[b]{.17\linewidth}
\centering
  \begin{tabular}{cc}
    \resizebox{1.0\linewidth}{!}{
  \includegraphics[width=\linewidth]{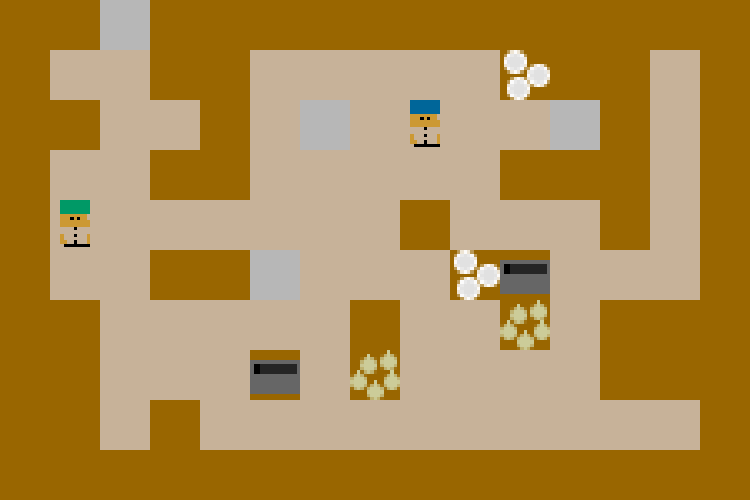}
   }   \end{tabular}
 \label{fig:plotT0}
\end{subfigure}
\vspace{-1cm}
\end{tabular}

\caption{Example Overcooked environments authored by different methods. The environments generated with the GAN+MIP approach are solvable by the human-robot team, while having design similarity to the human-authored environments.}
\label{fig:summary}
\end{figure*}

\section{Approach}

\noindent\textbf{Overview.} 
Our proposed framework consists of three main components: 1) A GAN which generates environments and is trained with human-authored examples. 2) A MIP which edits the generated environments to apply domain-specific constraints that make the environment solvable for the desired task. 3) A quality diversity algorithm, \mbox{CMA-ME}, which searches the latent space of the GAN to generate environments that maximize or minimize team performance, but are also diverse with respect to specified measures.


In Appendix~\ref{subsec:alternatives} we discuss two alternatives: optimizing latent space with random search instead of \mbox{CMA-ME}, and directly searching over the tiles of the environment, instead of exploring the latent space of a GAN.


\begin{figure}[t!]
\centering
\includegraphics[width=1.0\linewidth]{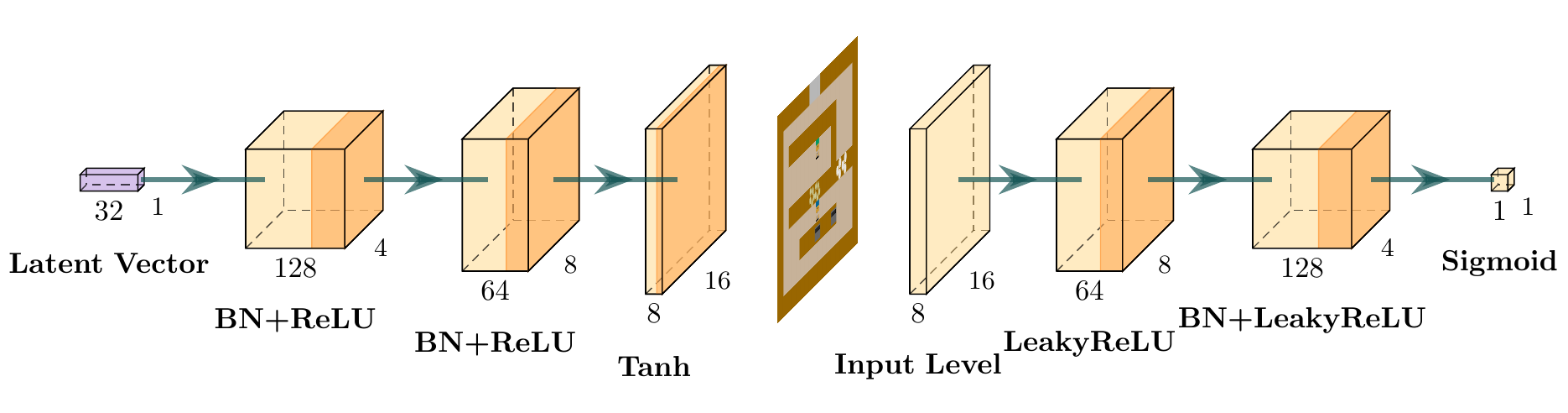}
\caption{Architecture of the GAN network.}
\label{fig:gan_arch}
\end{figure}

\noindent\textbf{Deep Convolutional GAN.} Directly searching over the space of possible environments can lead to the discovery of unrealistic environments. To promote realism, we incorporate GANs into our environment generation pipeline. Prior work~\cite{zhang2020video,fontaine2020illuminating} in PCG demonstrates that in the \textit{Super Mario Bros.} and \textit{Zelda} domains, GANs generate video game levels which adhere to the design characteristics of their training dataset.



We implemented a deep convolutional GAN (DCGAN) architecture identical to that in previous PCG work~\cite{zhang2020video,fontaine2020illuminating,volz2018evolving}, and we trained it on human-authored Overcooked environments. Fig.~\ref{fig:gan_arch} shows the architecture of the GAN and Fig.~\ref{fig:summary} shows example human-authored (top row) and GAN-generated (second row) environments. We provide implementation details in Appendix~\ref{subsec:gan-training}.

\noindent\textbf{Mixed-Integer Program Formulation.} While GAN-generated environments capture design similarities to their training dataset, most GAN-generated environments are unsolvable: GANs represent levels by assigning tile-types to locations, and this allows for walkable regions to become disconnected or item types to be missing or too frequent. In general, GANs struggle to constrain the generative environment space to satisfy logical properties~\citep{torrado:cog20}. 



To compensate for the limitations of GANs, we build upon the \textit{generate-and-repair} approach proposed in previous work~\cite{zhang2020video}, which repairs environments with mixed-integer linear programs (MIPs) that directly encode solvability constraints. To ensure that repairs do not deviate too far from the GAN-generated environment, the MIP minimizes the edit distance between the input and output environments. The result is an environment similar to the GAN-generated one that satisfies all solvability constraints.


In Overcooked, we specify MIP constraints that (1) bound the number of each tile type (i.e. starting agent locations, counters, stoves, floor), (2) ensure that key objects are reachable by both players and (3) prevent the agents from stepping outside the environment. We formalize the MIP in  Appendix~\ref{sec:MIP}.



The third row of Fig.~\ref{fig:summary} shows example environments generated by the GAN+MIP approach. We observe that the environments allow execution of the Overcooked game, while appearing similar to the GAN levels in the second row. In contrast, the fourth row (MIP-random) shows environments generated from random tile assignments passed as inputs to the MIP. While all environments are solvable, they are stylistically different than the human-authored examples: each environment appears cluttered and unorganized.


\noindent\textbf{Latent Space Illumination.}  
While the GAN generator and MIP repair collectively form a generative space of solvable environments, simply sampling the generative space is not efficient at generating diverse environments (see Appendix~\ref{sec:random-search}). 



We address this issue by formulating the problem as a Latent Space Illumination (LSI) problem~\citep{fontaine2020illuminating} defined below. Solving the LSI problem allow us to extract environments that are diverse with respect to fluency metrics while still maximizing or minimizing team performance.


\subsubsection{Problem Formulation} LSI formulates the problem of directly searching the latent space as a quality diversity (QD) problem. For quality, we provide a performance metric $f$ that measures team performance on the joint task,  e.g, time to completion or number of tasks completed. For diversity, we provide metric functions which measure how environments should vary, e.g. the distribution of human and robot workloads. These metrics, defined in QD as behavior characteristics (BCs), form a Cartesian space known as a behavior space. In the MAP-Elites~\citep{cully:nature15, mouret2015illuminating} family of QD algorithms, the behavior space is partitioned into $N$ cells to form an archive of environments. Visualizing the archive as a heatmap allows researchers to interpret how performance varies across environments inducing different agent behaviors.


LSI searches directly for GAN latent codes $\bm{z}$. After simulating the agents on the generated environment, each code $\bm{z}$ maps directly to a performance value $f(\bm{z})$ and a vector of behaviors $\bm{b}(\bm{z})$. We discuss different performance metrics and BCs in section~\ref{sec:environments}.



The objective of LSI is to maximize the sum of expected performance values $f$:

\begin{equation}
  M(\bm{z_1}, ... ,\bm{z_N}) =  \max \sum_{i=1}^N \mathbb{E}[f(\bm{z_i})]
\label{eq:objective}
\end{equation}

In Eq.~\ref{eq:objective}, $\bm{z_i}$ refers to the latent vector occupying cell $i$. We note that the human and robot policies and the environment may be stochastic, therefore we estimate the expected performance over multiple trial runs. 


\subsubsection{CMA-ME for Latent Space Illumination} 
We choose Covariance Matrix Adapation \mbox{MAP-Elites} (\mbox{CMA-ME})~\citep{fontaine2019covariance}, a state-of-the-art QD algorithm, to solve the LSI problem. \mbox{CMA-ME} outperformed other quality diversity algorithms when illuminating the latent space of a GAN trained to generate \textit{Super Mario Bros.} levels~\citep{fontaine2020illuminating}.


\mbox{CMA-ME} combines MAP-Elites with the adaptation mechanisms of \mbox{CMA-ES}~\cite{hansen:cma16}. To generate new environments, latent codes $\bm{z}$ are sampled from a Gaussian $\mathcal{N}(\bm{\mu}, C)$ where each latent code $\bm{z}$ corresponds to an environment. After generating and repairing the environment with the GAN and MIP, we simulate the agents in the environment and compute agent performance $f$ and behavior characteristics $\bm{b}$. The behavior $\bm{b}$ then maps to a unique cell in the archive. We compare our new generated environment with the existing environment of that cell and replace the environment if the new environment has a better $f$ value. The distribution $\mathcal{N}(\bm{\mu}, C)$ is finally updated based on how the archive has changed, so that it moves towards underexplored areas of behavior space. We provide the pseudocode for CMA-ME adapted to solve the LSI problem in Appendix~\ref{sec:LSI-CMAME}.



\section{Planning Algorithms}

We consider two planning paradigms: (1) where human and robot follow a centralized joint plan and (2) where the robot reasons over the partially observable human subtask.

\label{sec:algorithms}
\noindent\textbf{Centralized Planning.} 
\label{subsec:centralized}
Here human and robot follow a centralized plan specified in the beginning of the task. We incorporate the near-optimal joint planner of previous work~\cite{carroll2019utility}, which pre-computes optimal joint motion plans for every possible start and goal location of the agents, and optimizes the motion plan costs with an $A^*$ planning algorithm~\citep{hart1968formal}. 


\noindent\textbf{Human-Aware Planning.} \label{subsec:human-aware_planning}
\subsubsection{Robot Planner.}
We examine the case where the robot is not aware of the subtask the human is currently aiming to complete, e.g., picking up an onion. We model the robot as a QMDP planner, wherein the human subtask is a partially observable variable.

To make the computation feasible in real-time, we only use the QMDP to decide subtasks, rather than low-level actions. To move to the location of each subtask, the robot follows a  motion plan precomputed in the same way as in the centralized planning. The motion plan assumes that both agents move optimally towards their respective subtasks.  


The QMDP planner \textit{adapts} to the human: it observes the human low-level motions to update its belief over the human subtasks, and selects a subtask for the robot that minimizes the expected cost to go. We describe our QMDP implementation in Appendix~\ref{sec:qmdp_implementation}.

\subsubsection{Human Planner} We selected a rule-based human model from previous work~\cite{carroll2019utility}, which myopically selects the highest priority subtask based on the world state. The model does not reason over an horizon of subtasks and does not account for the robot's actions. Empirically, we found the model to perform adequately most of the time, when users choose actions quickly.


\section{Environments} \label{sec:environments}
We performed \num{4} different experiments to demonstrate that our proposed framework generates a variety of environments that result in a diverse set of coordination behaviors.\footnote{Videos of task executions of all environments in the figures of this section are included in \url{https://github.com/icaros-usc/overcooked_lsi_videos}.}

In all experiments we use the same performance metric $f$, which is a function of both the number of completed orders and the amount of time each order was completed. We describe the metric and other implementation details in Appendix~\ref{sec:implementation}. 


\noindent\textbf{Workload Distributions with Centralized Planning.}
We generate environments that result in a broad spectrum of workload distributions, such as environments where only one agent does all the work and environments where both agents contribute evenly to the task. Both agents execute a precomputed centralized joint plan.

To assess differences in workloads, we specify as BCs the differences (robot minus human) in the number of actions performed for each object type: number of ingredients (onions) held, number of plates held, and number of orders delivered. 

\begin{figure}
\centering
\begin{subfigure}{0.5\textwidth}
    \includegraphics[width=\linewidth]{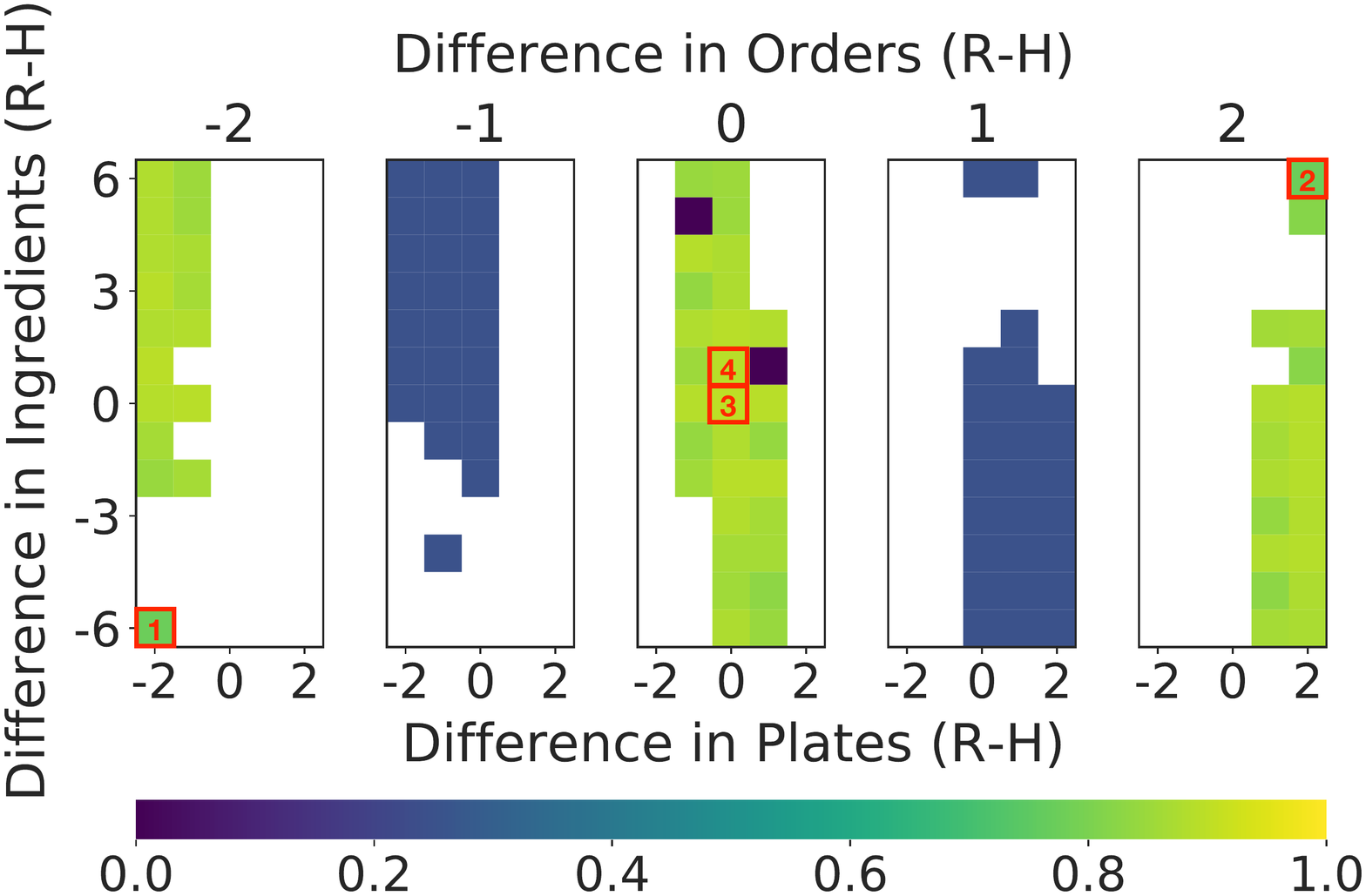}
    \label{fig:my_label}
\vspace{-1em}
\end{subfigure}\hfil
\begin{tabular}{cc}
\begin{subfigure}{0.2\textwidth}
    \includegraphics[width=\linewidth]{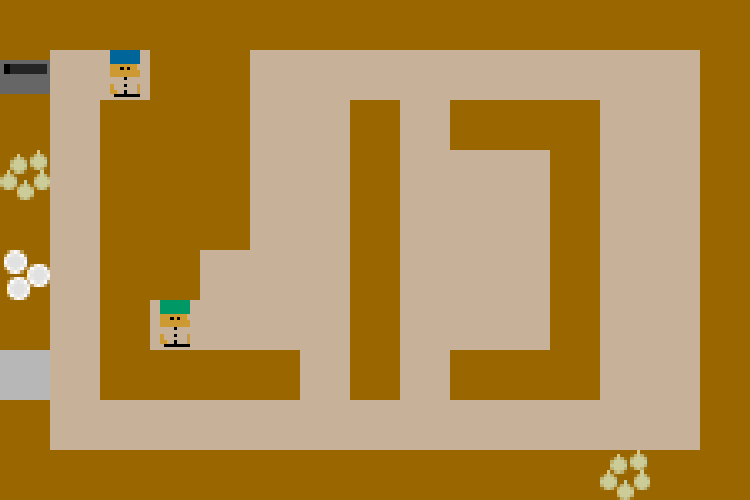}
    \caption*{(1)}
    \label{fig:my_label}
\end{subfigure}&
\begin{subfigure}{0.2\textwidth}
    \includegraphics[width=\linewidth]{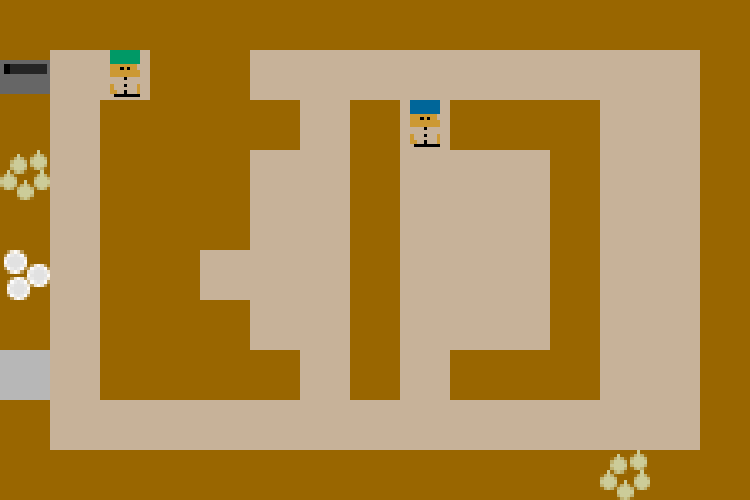}
    \caption*{(2)}
    \label{fig:my_label}
\end{subfigure}\\
\begin{subfigure}{0.2\textwidth}
    \includegraphics[width=\linewidth]{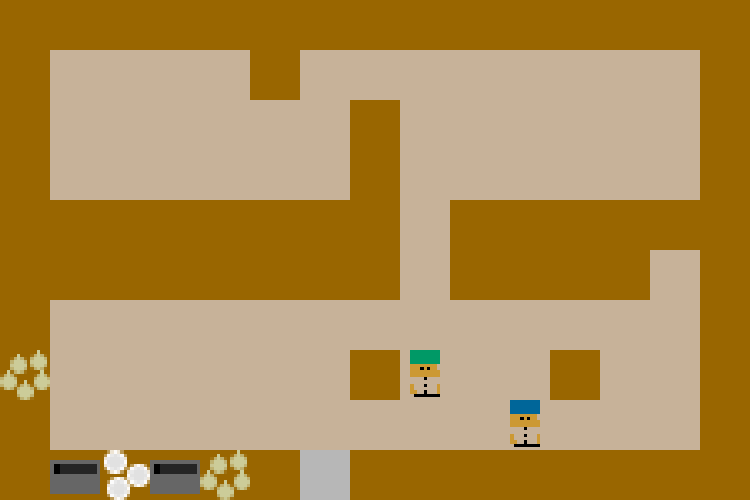}
    \caption*{(3)}
    \label{fig:my_label}
\end{subfigure}&
\begin{subfigure}{0.2\textwidth}
    \includegraphics[width=\linewidth]{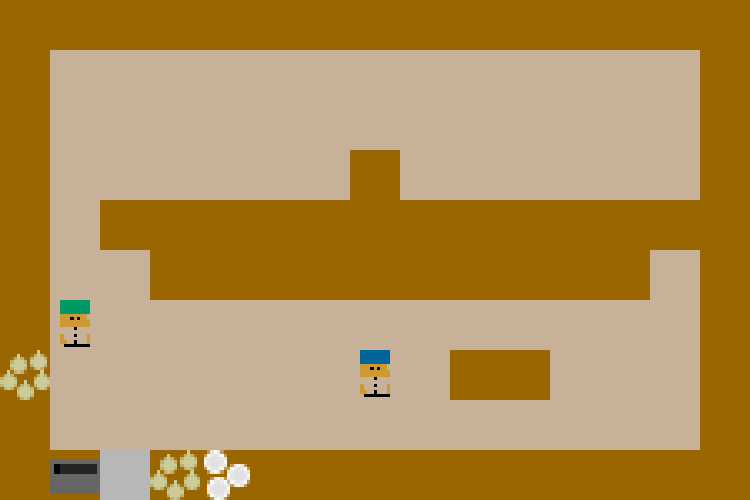}
    \caption*{(4)}
    \label{fig:my_label}
\end{subfigure}\\
\end{tabular}
\caption{Archive of environments with different workload distributions for the centralized planning agents and four example environments corresponding to different cells in the archive. Environments (1,2) resulted in uneven workload distributions, while environments (3,4) resulted in even workload distributions. We annotate four environments from the archive. The bar shows the normalized value of the objective $f$.} 
\label{fig:centralized}
\end{figure}

Fig.~\ref{fig:centralized} shows the generated archive: we illustrate the 3D behavior space as a series of five 2D spaces, one for each value of the difference in orders. Each colored cell represents an environment with BCs computed by simulating the two agents in that environment. Lighter colors indicate higher performance $f$. 

We observe that when the difference in orders is \num{-1} or \num{+1}, performance is low; this is expected since there are only 2 orders to deliver, thus an absolute difference of \num{+1} means that only one order was delivered. 

We simulate the two agents in environments of the extreme regions of the archive to inspect how these environments affect the resulting workload distributions. For instance, either the robot (green agent) or the simulated human (blue agent) did all of the work in environments (1) and (2) of Fig.~\ref{fig:centralized} respectively. 
We observe that in these environments the dish dispenser, onion dispenser, cooking pot and serving counter are aligned in a narrow corridor. The optimal joint plan is, indeed, that the agent inside the corridor picks up the onions, places them in the pot, picks up the plate and delivers the onions. 

On the other hand, in environments (3) and (4), the workload was distributed exactly or almost exactly evenly. We see that in these environments, both agents have easy access to the objects. Additionally, all objects are placed next to each other. This is intentional, since \mbox{CMA-ME} attempts to fill the archive with diverse environments that each maximize the specified performance metric. Since performance is higher when all orders are delivered in the minimum amount of time, positioning the objects next to each other results in shorter time to completion. 

This object configuration works well in centralized planning since the agents precompute their actions in advance, and there are no issues from lack of coordination. We observe that this is not the case in the human-aware planning experiments below.



\label{subsec:human-aware}
\noindent\textbf{Workload Distributions with Human-Aware Planning.} In this experiment, the human and robot do not execute a precomputed centralized joint plan. Instead, the robot executes a QMDP policy and the human executes a myopic policy. 


We run two experiments: In the first experiment, we generate environments that maximize the performance metric $f$, identical to Section~\ref{subsec:centralized}. In the second, we attempt to find environments that \textit{minimize} the performance metric. The latter is useful when searching for failure cases of developed algorithms~\cite{fontaine2020quality}. We are specifically interested in drops in performance  that arise from the assumptions of the QMDP formulation, rather than, for example, poor performance because objects are too far from each other. Therefore, for the second experiment we use as a baseline the performance of the team when the robot executes an MDP policy that fully observes the human subtask, and we \textit{maximize the difference} in performance between simulations with the MDP policy and the QMDP policy.

We note that in decentralized planning, the two agents may get ``stuck'' trying to reach the same object. We adopt a rule-based mechanism from previous work~\cite{carroll2019utility} that selects random actions for both agents until they get unstuck. While the MDP, QMDP, and myopic human policies are deterministic, the outcomes are stochastic because of these random actions. Therefore, we run multiple trials in the same environment, and we empirically estimate the performance and BCs with their median values (see Appendix~\ref{subsec:trials}). 



\begin{figure}
\centering
\begin{subfigure}{0.5\textwidth}
    \includegraphics[width=\linewidth]{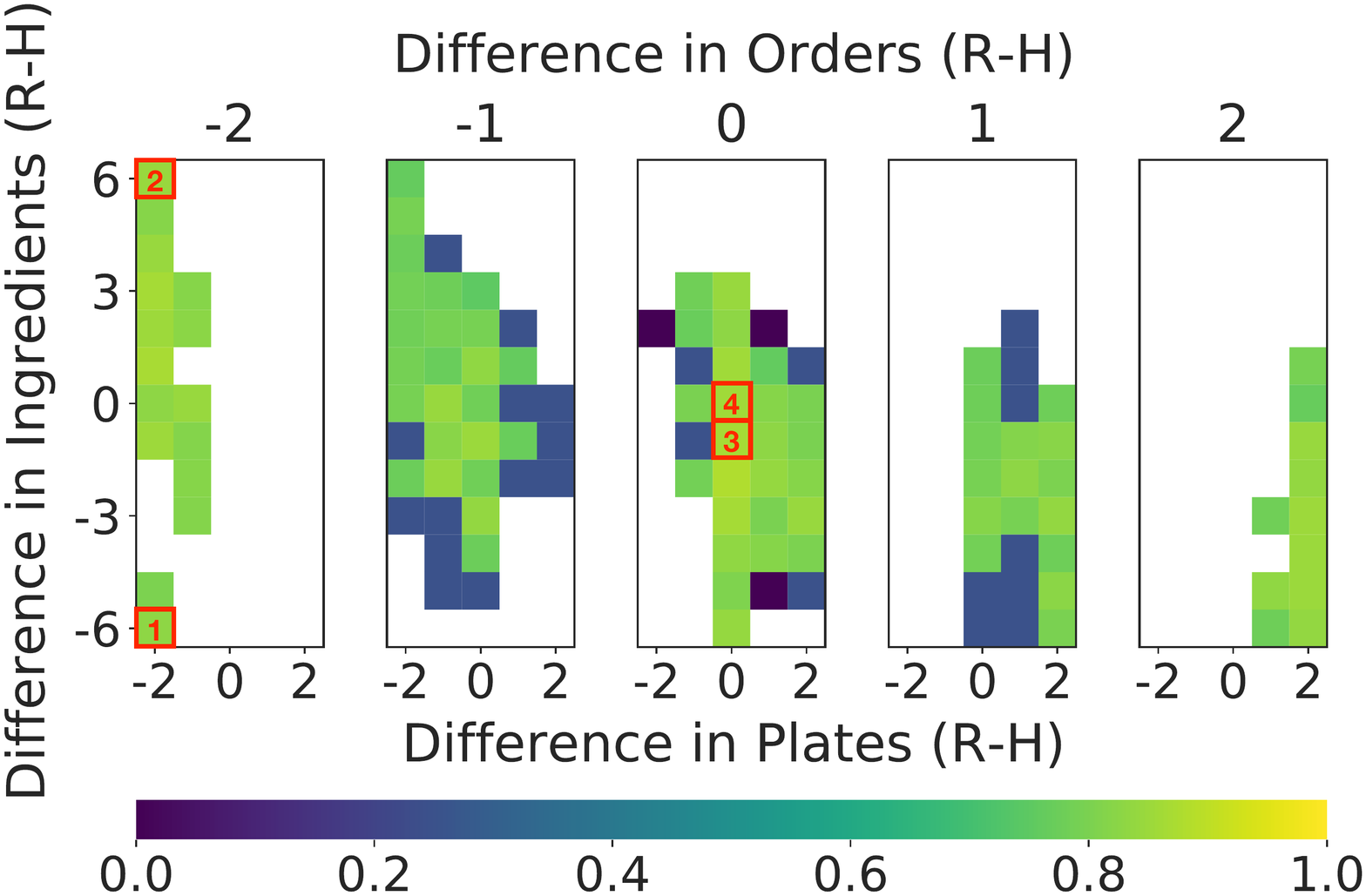}
    \label{fig:qmdp_max_map}
\vspace{-1em}
\end{subfigure}\hfil
\begin{tabular}{cc}
\begin{subfigure}{0.2\textwidth}
    \includegraphics[width=\linewidth]{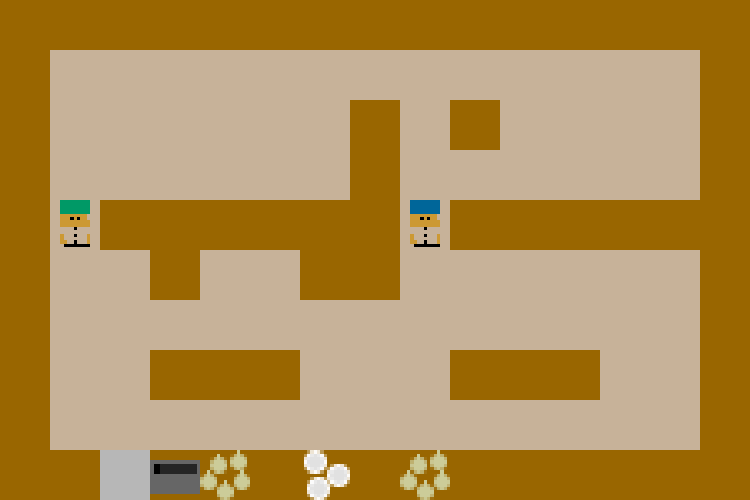}
    \caption*{(1)}
    \label{fig:qmdp_max_map_1}
\end{subfigure}&
\begin{subfigure}{0.2\textwidth}
    \includegraphics[width=\linewidth]{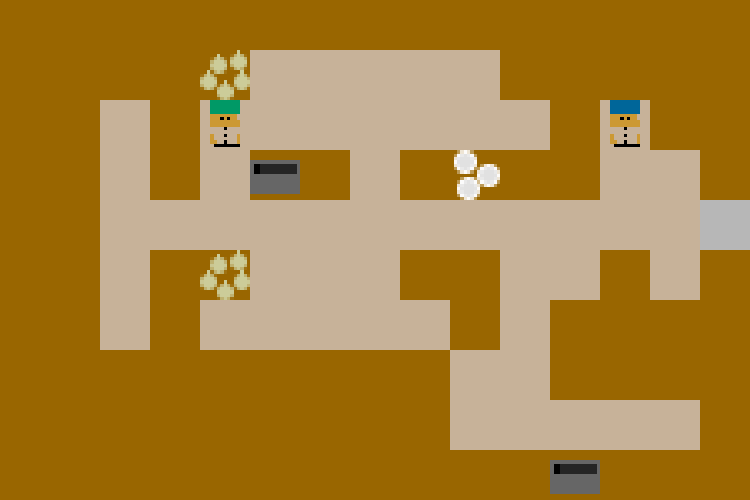}
    \caption*{(2)}
    \label{fig:qmdp_max_map_2}
\end{subfigure}\\
\begin{subfigure}{0.2\textwidth}
    \includegraphics[width=\linewidth]{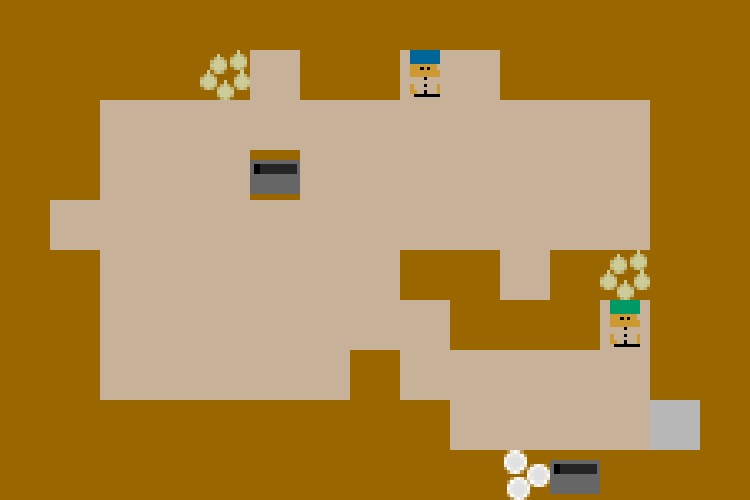}
    \caption*{(3)}
    \label{fig:qmdp_max_map_3}
\end{subfigure}&
\begin{subfigure}{0.2\textwidth}
    \includegraphics[width=\linewidth]{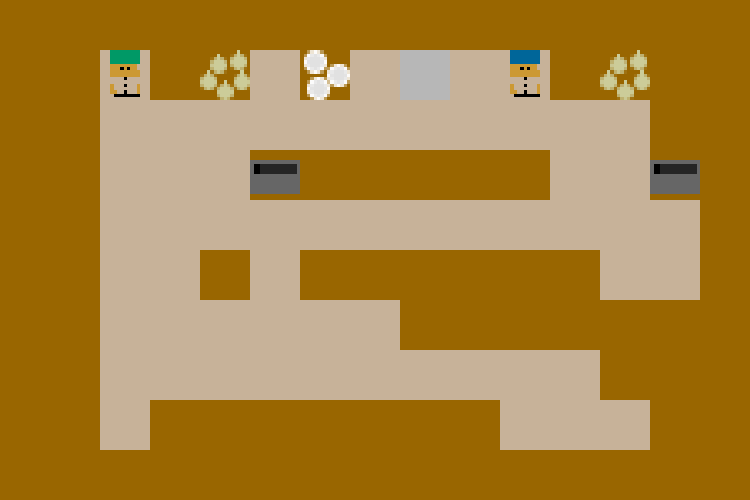}
    \caption*{(4)}
    \label{fig:qmdp_max_map_4}
\end{subfigure}\\
\end{tabular}
\caption{Archive of environments with different workload distributions of a QMDP robot and a simulated myopic  human.}
\label{fig:qmdp_max_map}
\end{figure}


\subsubsection{Maximizing Performance}
Inspecting the environments in the extremes of the generated archive (Fig.~\ref{fig:qmdp_max_map}) reveals interesting object layouts. For example, all objects in environment (1) are aligned next to each other, and the agent that gets first in front of the leftmost onion dispenser ``blocks'' the path to the other agent and completes all the tasks on its own. In environment (2), the robot starts the task next to the onion dispenser and above the pot. Hence, it does all the onion deliveries, while the human picks up the onions and delivers the order by moving below the pot. 

In environments (3) and (4),  each agent can do the task independently, because each team member is close to their own pot. This results in even workload distribution. The two agents achieve high performance, since no delays arise from lack of coordination. 




\subsubsection{Minimizing Performance} \label{subsec:minimizing} 
\begin{figure}
\centering
\begin{subfigure}{0.5\textwidth}
    \includegraphics[width=\linewidth]{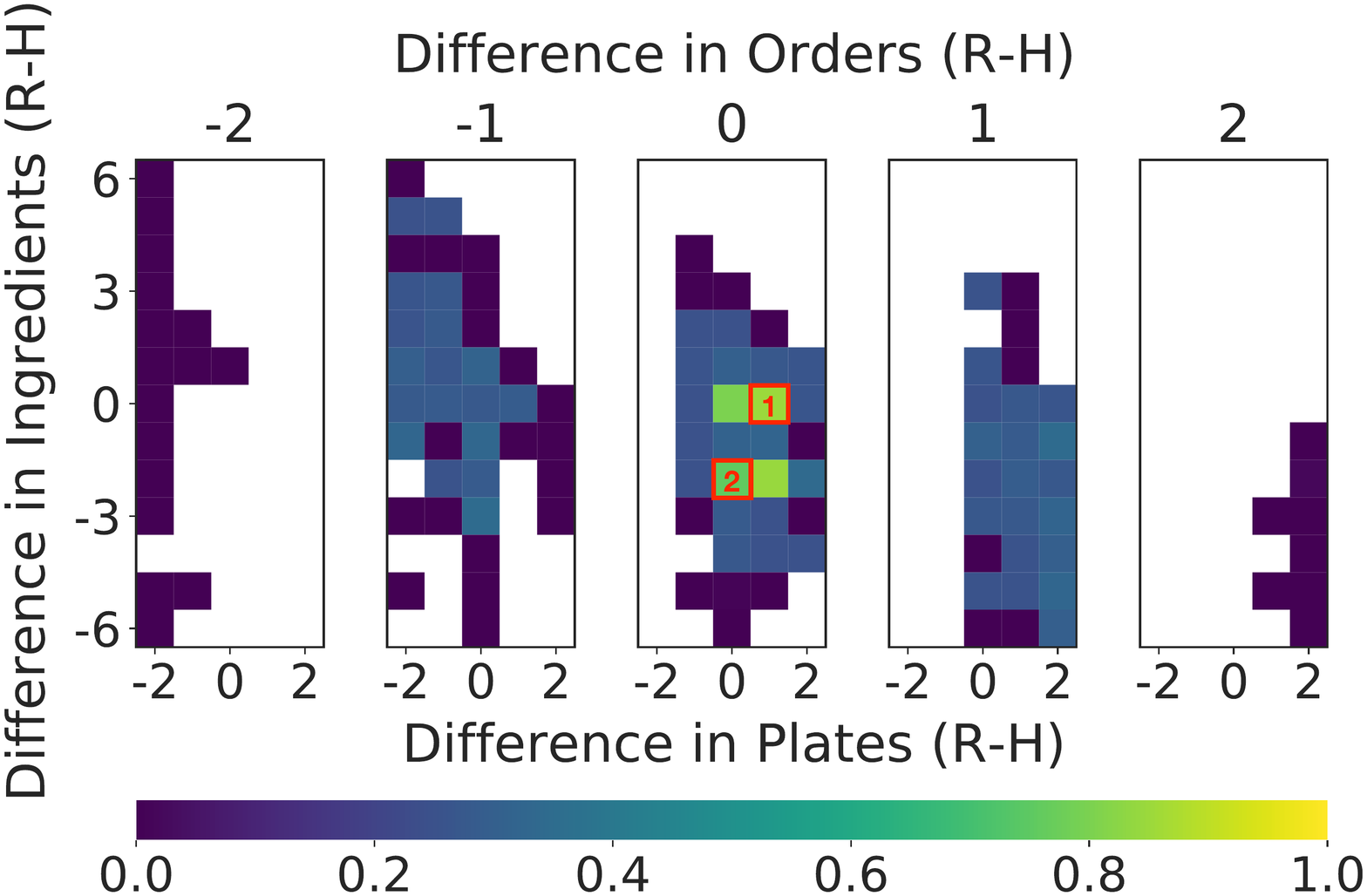}
\vspace{-1em}
\end{subfigure}\hfil
\begin{tabular}{cc}
\begin{subfigure}{0.2\textwidth}
    \includegraphics[width=\linewidth]{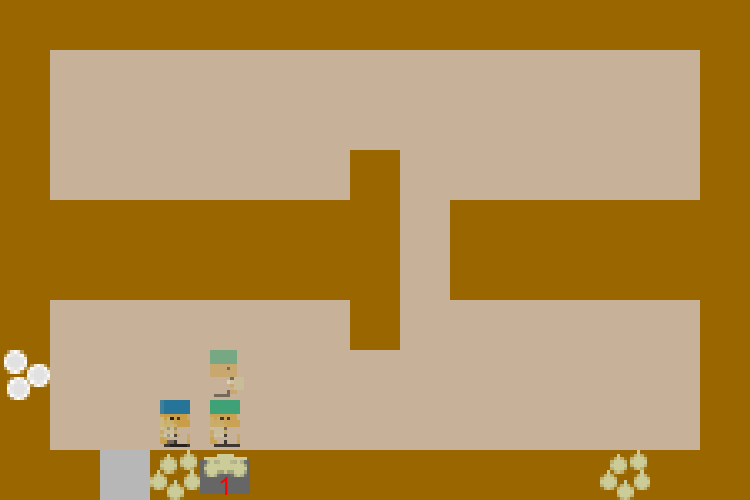}
   \caption*{(1a)}
    \label{fig:qmdp_min_map_1_0}
\end{subfigure}&
\begin{subfigure}{0.2\textwidth}
    \includegraphics[width=\linewidth]{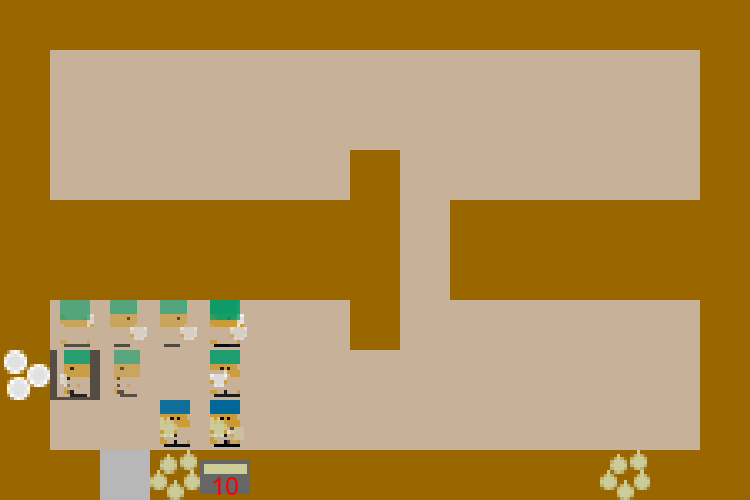}
   \caption*{(1b)}
    \label{fig:qmdp_min_map_1_1}
\end{subfigure}\hfil
\end{tabular}
\begin{tabular}{cc}
\begin{subfigure}{0.2\textwidth}
    \includegraphics[width=\linewidth]{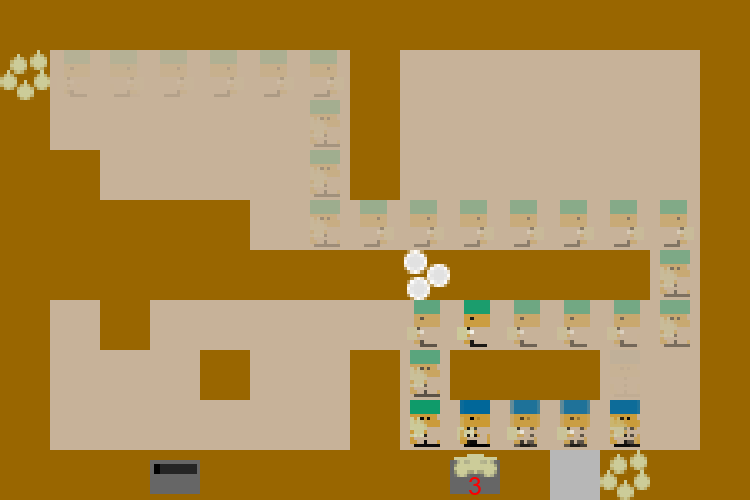}
    \caption*{(2a)}
    \label{fig:qmdp_min_map_2_0}
\end{subfigure}&
\begin{subfigure}{0.2\textwidth}
    \includegraphics[width=\linewidth]{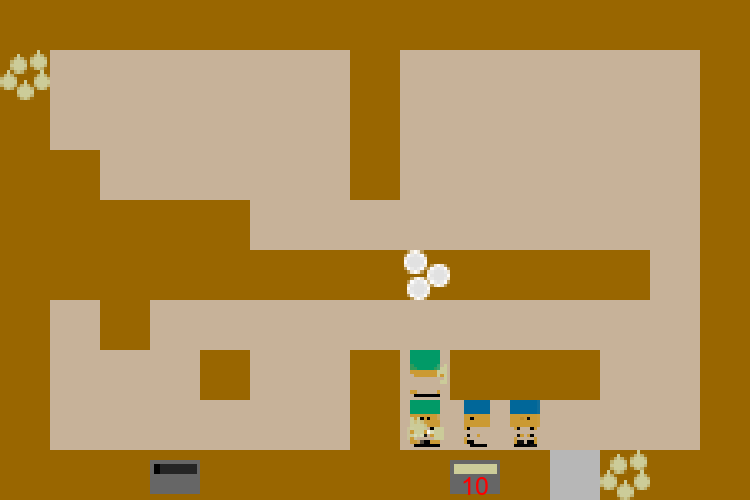}
    \caption*{(2b)}
    \label{fig:qmdp_min_map_2_1}
\end{subfigure}\\
\end{tabular}
\caption{Archive of environments when attempting to \textit{minimize} the performance of a QMDP robot (green agent) and a simulated myopic human (blue agent). Lighter color indicates lower performance. (1a) and (1b) show successive frame sequences for environment (1), and similarly (2a), (2b) for environment (2) (see videos in supplemental material).}
\label{fig:qmdp_min_map}
\end{figure}

In the generated archive of Fig.~\ref{fig:qmdp_min_map}, lighter colors indicate \textit{lower} performance of the team of the QMDP robot and the myopic human compared to an MDP robot and a myopic human. We are particularly interested in the environments where the team fails to complete the task. 


In environment (1) of Fig.~\ref{fig:qmdp_min_map}, the simulated human picks up an onion at exactly the same time step that the robot delivers the third onion to the pot. There is now no empty pot to deliver the onion, so the human defaults to going to the pot and waiting there, blocking the path of the robot. The environment leads to an interesting edge case that was not accounted for in the hand-designed human model but is revealed by attempting to minimize the performance of the agents.

In environment (2) of Fig.~\ref{fig:qmdp_min_map}, the two agents get stuck in the narrow corridor in front of the rightmost onion dispenser. Due to the ``auto-unstuck'' mechanism, the simulated human goes backward towards the onion dispenser. The QMDP planner, which uses the change of distance to the subtask goal location as observation (see Appendix~\ref{sec:qmdp_implementation}), erroneously infers that the human subtask is to reach the onion dispenser, and does not move backwards to allow the human to go to the dish dispenser. This environment highlights a limitation of the distance-based observation function since it is not robust to random motions that occur when the two agents get stuck. 

Overall, \emph{we observe that when minimizing performance, the generated environments reveal edge cases that can help a designer better understand, debug, and improve the agent models.}

\noindent\textbf{Team Fluency with Human-Aware Planning.} An important aspect  of  the  quality of the interaction between two agents is their team fluency. One team fluency metric is the \textit{concurrent motion} (also defined as concurrent activity), which is defined in~\cite{hoffman2019evaluating} as ``the percentage of time out of the total task time, during which both agents have been active concurrently.''

We include as second metric the number of time steps the agents are ``stuck,'' which occurs when both agents are in the same position and orientation for two successive time steps. We use the human-aware planning models as in the previous experiment, and we search for environments that maximize team performance but are diverse with respect to the concurrent motion and time stuck of the agents.


We observe in the generated archive (Fig.~\ref{fig:fluency_map}) that environments with higher concurrent motion have better performance, since the agents did not spend much time waiting.  For example, environments (3) and (4) result in very high team fluency. These environments have two pots and two onion dispensers, which are easily accessible. 


On the other hand, example environments (1) and (2) have poor team fluency. These environments have long corridors, and one agent needs to wait a long time for the second agent to get out of the corridor. 
In environment (1), the two agents get stuck  when the myopic human attempts to head towards the onion dispenser ignoring the robot, while the QMDP agent incorrectly assumes that the human gives way to the robot.






\begin{figure}
\centering
\begin{subfigure}{0.33\textwidth}
    \includegraphics[width=\linewidth]{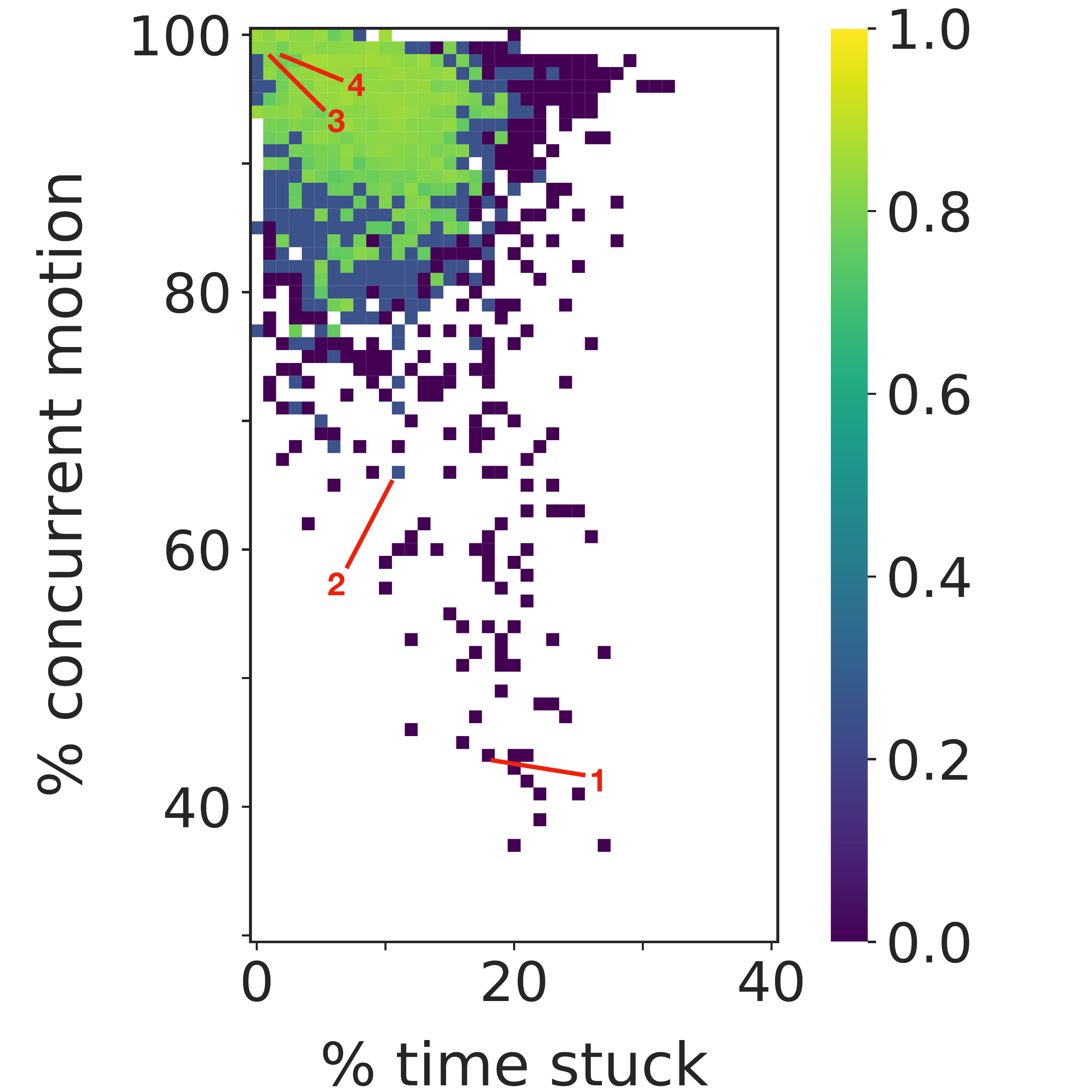}
    \label{fig:fluency_map}
\end{subfigure}\hfil
\begin{tabular}{cc}
\begin{subfigure}{0.2\textwidth}
    \includegraphics[width=\linewidth]{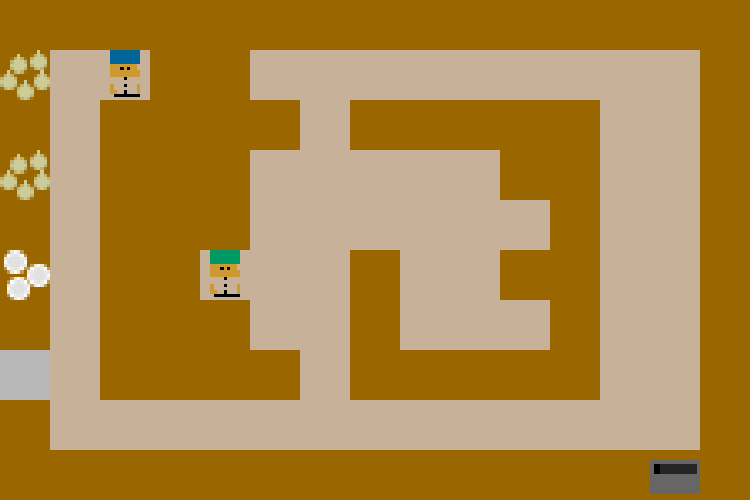}
    \caption*{(1)}
    \label{fig:fluency_map_1}
\end{subfigure}&
\begin{subfigure}{0.2\textwidth}
    \includegraphics[width=\linewidth]{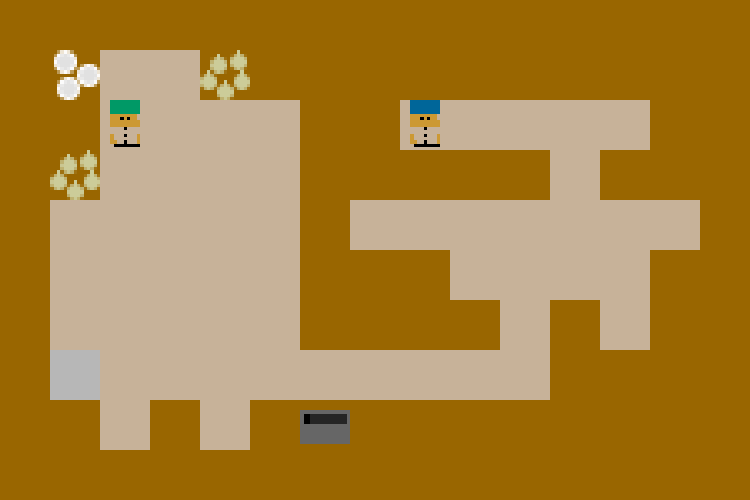}
    \caption*{(2)}
    \label{fig:fluency_map_2}
\end{subfigure}\\
\begin{subfigure}{0.2\textwidth}
    \includegraphics[width=\linewidth]{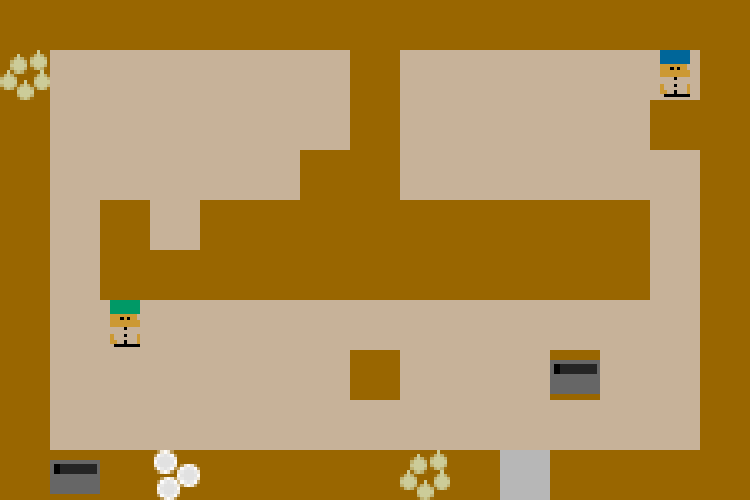}
    \caption*{(3)}
    \label{fig:fluency_map_3}
\end{subfigure}&
\begin{subfigure}{0.2\textwidth}
    \includegraphics[width=\linewidth]{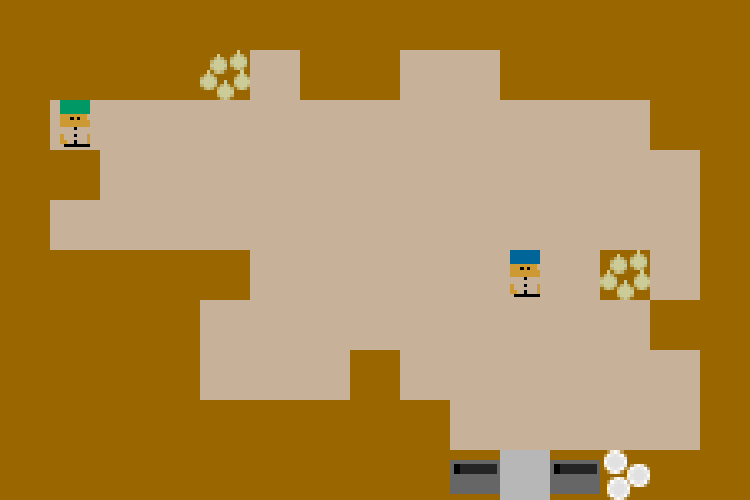}
    \caption*{(4)}
    \label{fig:fluency_map_4}
\end{subfigure}\\
\end{tabular}
\caption{Archive of environments with different team fluency metrics. Environments (1) and (2) resulted in low team fluency, while (3) and (4) resulted in high team fluency.}
\label{fig:fluency_map}
\end{figure}


\begin{table*}[t]
\centering
\begin{tabular}{l|ccc|ccc|ccc}
\hline
             & \multicolumn{3}{l|}{Workload  Distribution Max}          & \multicolumn{3}{l|}{Workload  Distribution Min} & \multicolumn{2}{l}{Team Fluency}  \ \\ 
    \toprule
$\epsilon$    & Diff. in Ingredients & Diff. in Plates & Diff. in Orders & Diff. in Ingredients & Diff. in Plates & Diff. in Orders & Concurrent Motion & Stuck \\
    \midrule
0.00 &0.85 &  0.67 & 0.66 & 0.85& 0.78 &0.76 & 0.88  & 0.92\\
0.05  &0.76 & 0.55 & 0.50 & 0.79 & 0.63 &  0.61 & 0.77 & 0.85 \\
0.10   & 0.68 & 0.46& 0.39 & 0.72 &  0.52 & 0.50 & 0.67 & 0.76 \\
0.20    & 0.56  & 0.33 & 0.23 & 0.63 & 0.40 & 0.35 & 0.52 &  0.62 \\
0.50 & 0.46 & 0.22 & 0.08 & 0.39 & 0.26 &   0.11 &  0.34 & 0.30 \\
  \bottomrule
\end{tabular}
\caption{Spearman's rank-order correlation coefficients between the computed BCs and the initial placement of environments in the archive for increasing levels of noise $\epsilon$ in human inputs.}
\label{tab:results}
\end{table*}

\section{Robustness to Human Model} \label{sec:robustness}
When evaluating the generated environments in human-aware planning, we assumed a myopic human model. We wish to test the robustness of the associated coordination behaviors with respect to the model: if we add noise, will the position of the environments in the archive change? 





Therefore, for each cell in the archives from Fig.~\ref{fig:qmdp_max_map} and~\ref{fig:qmdp_min_map}, and for 100 randomly selected cells in the archive of Fig.~\ref{fig:fluency_map}, we compute the BCs for increasing levels of noise in the human actions. We simulate noisy humans by using an $\epsilon$-myopic human model, where the human follows the myopic model with a probability of $1-\epsilon$ and takes a uniformly random action otherwise. We then compute the Spearman's rank-order correlation coefficient between the initial position of each environment in the archive and the new position, specified by the computed BCs, in the presence of noise.





Table~\ref{tab:results} shows the computed correlation coefficients for each BC and for increasing values of $\epsilon$. The ``Workload Distribution Max,'' ``Workload Distribution Min,'' and ``Team Fluency'' refer to the archives of Fig.~\ref{fig:qmdp_max_map},~\ref{fig:qmdp_min_map} and Fig.~\ref{fig:fluency_map}. All values are statistically significant with Bonferroni correction ($p<0.001$). 

We observe that even when $\epsilon = 0$, the correlation is strong but not perfect, since there is randomness in the computed BCs because of the ``auto-stuck'' mechanism. Values of $\epsilon=0.05$ and $0.1$ result in moderate correlation between the initial and new position of the environments in the archive. The correlation appears to be stronger for the difference in ingredients. This is because the environments with extreme values of this BC ($+6, -6$) had nearly zero variance since one agent would consistently ``block'' the other agent from accessing the onion dispenser. As expected, when the simulated human becomes random \SI{50}{\%} of the time ($\epsilon = 0.5$), there is only a weak, albeit still significant, correlation.

\section{User Study}
Equipped with the findings from section~\ref{sec:robustness}, we want to assess whether the differences in coordination observed in simulation translate to actual differences when the simulated robot interacts with real users. 

We selected \num{12} environments from the generated archives in the human-aware planning experiments of section~\ref{sec:environments}, including \num{3} ``even workload'' environments, \num{3} ``uneven workload'' environments'', \num{3} ``high team fluency'' environments and \num{3} ``low team fluency'' environments.  These environments are shown in Appendix~\ref{sec:study2}.

\noindent\textbf{Procedure.} Participants conducted the study remotely by logging into a server while video conferencing with the experimenter. Users controlled the human agent with the keyboard and interacted with the QMDP robot. The experimenter first instructed them in the task and asked them to complete three training sessions, where in the first two they practiced controlling the human with their keyboard and in the third they practiced collaborating with the robot. They then performed the task in all \num{12} environments in randomized order (within subjects design). We asked all participants to complete the tasks as quickly as possible.\footnote{All the anonymized user data and instructions about visualizing all the playthroughs are in \url{https://github.com/icaros-usc/overcooked_lsi_user_study}.}

\noindent\textbf{Participants.} We recruited \num{27} participants from the local community  (ages 20-30, M=23.81, SD=2.21). Each participant was compensated $\$ 8$ for completing the study, which lasted approximately \num{20} minutes. 

\noindent\textbf{Hypotheses.}

\noindent{H1.} The difference in the workloads between the human and the robot will be larger in the ``uneven workload'' environments, compared to the ``even workload'' environments.

\noindent{H2.} The team fluency of the human-robot team will be better in the ``high team fluency'' environments, compared to the ``low team fluency'' environments.

\noindent\textbf{Dependent Measures.} Identically to the experiments in section~\ref{sec:environments}, we computed the following BCs: the difference in the ingredients, plates, and orders from the playthroughs in the ``even/uneven workload'' environments, and the percentage of concurrent motion and time stuck in the ``low/high team fluency'' environments.

We used the average over the BCs computed from the three environments of the same type, e.g., the three ``even workload'' environments, as an aggregate measure. For the analysis, we used the absolute values of the workload differences, since we are interested in whether the workload is even or uneven and not which agent performed most of the actions.  


\noindent\textbf{Analysis.} A Wilcoxon signed-rank test determined that two out of the three workload BCs  (difference in ingredients: $z = 3.968, p < 0.001$, difference in orders: $z = 2.568, p=0.01)$ were significantly larger in the ``uneven workload'', compared to the ``even workload'' environments.

Additionally, a Wilcoxon signed-rank test showed that the percentage of concurrent motion was significantly higher for the high team fluency environments, compared to the low team fluency ones ($z=4.541,p<0.001$). There were no  significant differences in the time stuck, since users resolved stuck situations quickly by giving way to the robot.

These results support our hypotheses and show that the differences in the coordination observed in simulation translate to differences observed in interactions with actual users. We plot histograms of the computed BCs and discuss participants' open-ended responses  in Appendix~\ref{sec:study2}.


\section{Discussion}
\noindent\textbf{Limitations.} Our work is limited in many ways. 
Our user study was conducted online with a simulated robot. Our future goal is to evaluate our framework on human-robot experiments in a real-world collaborative cooking setting, where users are exposed to different scenes. Further experiments with a variety of robot and human models would expand the diversity of the generated environments and the observed behaviors. Another limitation is that, while the framework is agnostic of the agents' models, our framework requires human input for specifying the behavior characteristics and MIP constraints. Automating part of the solvability specification is an exciting area for future work.




\noindent\textbf{Implications.} We envision our framework as a method to help evaluate human-robot coordination in the future, as well as a reliable tool to help practitioners debug or tune their coordination algorithms. More generally, our framework can facilitate understanding of complex human-aware algorithms executing in complex environments. We are excited about future work that highlights diverse behaviors in different settings where coordination is essential, such as  manufacturing and assistive care. Finally, we hope that our work will guide future human-robot coordination research to consider the environment as a significant factor in coordination problems.


\bibliographystyle{plainnat}
\bibliography{references}

\appendices

\section{Mixed-Integer Linear Program Repair} \label{sec:MIP}

We adapt the problem formulation from \citep{zhang2020video} for repairing \textit{The Legend of Zelda} levels to \textit{Overcooked}. Since the \textit{Overcooked} environments require different constraints to guarantee environment solvability for the human-robot team, our exact formulation differs in the type of objects and domain-specific constraints. For replicability and completeness we provide the entire MIP formulation.

To generate solveable environments, we first formulate the \textit{space graph}~\citep{togelius2014procedural} that governs agent motion in possible environments. Let $G=(V, E)$ be a directed space graph were $V$ is the vertex set and $E$ is the edge set. Each vertex in $V$ represents a \emph{location} an object will occupy. Now consider the movement dynamics of each agent (human or robot) where each agent can move up, down, left, or right. Each edge $(i,j) \in E$ represents possible motion between location $i$ and location $j$ for an agent if no object impedes their motion.

To generate an environment, we solve a matching problem between object types $O$ and locations $V$. In the simplified \textit{Overcooked AI} environment there are 8 object types and $15 \times 10 = 150$ different tile locations. If unconstrained further, there are $8^{150} \approx 2.9 \cdot 10^{135}$ different environments possible. Object types $O$ include the human $h$, \mbox{the robot $r$}, \mbox{countertops $c$}, empty space (floor) $e$, serve points $s$, dish dispensers $d$, onion dispensers $n$, and pots (with stoves) $p$.

To formulate the matching in the MIP, we create a vector of binary decision variables for each pair of object type $o \in O$ and location $v \in V$ in the space graph. For example, if variable $s_v$ were assigned to $1$, then vertex $v$ in the space graph would contain a serve point. Assigning $s_v$ to $0$ means that vertex $v$ does not contain a serve point. Finally, we constrain each vertex to contain exactly one object type:

\setlength\abovedisplayskip{0pt}
\setlength\belowdisplayskip{10pt}
\begin{equation} \label{eq:node_unique}
    h_v + r_v + c_v + e_v + s_v + d_v + n_v + p_v = 1, 
    \forall v \in V
\end{equation}

\noindent\textbf{Solvability Constraints.} While the above formulation ensures that a feasible solution to the MIP results in an \textit{Overcooked} environment, the generative space of environments must be further constrained to ensure each generated environment is solveable by the human-robot team. 

Importantly, agents must be able to reach key objects in the environment. For example, the human must have an unobstructed path to the stove, dish dispenser, ingredients, and serve point. 

We model the reachability problem as a flow problem~\cite{Goldberg2014flow}. Network flow can be modelled as a linear program, and as a result we can incorporate flow constraints into our MIP formulation. To model flow, we create non-negative integer variables $f(u,v) \in \mathbb{Z}_{\geq0}$ for each edge $e=(u,v) \in E$ in the space graph $G$.

Now consider special object types $S \subseteq O$, for source object types, and $T \subseteq O$, for sink object types. We require that a path from a source object exists to each sink object. Specifically, we require that a path exists from the human $h$ to all empty space, serve points, dish dispensers, onion dispensers, pots, and the robot by setting $S = \{h\}$ and $T = \{e, s, d, n, p, r\}$. Note that if we allow the human to reach the robot, then the robot can reach all other objects in $T$. \footnote{This holds only if the free space does not form a line graph, but empirically we found this constraint to be sufficient for large environments}

However, we must also require that the path is unobstructed. Therefore, each path must not pass through countertops or other objects that impede movement. Let $B \subseteq O$ be the set of all object types that can impede movement. In \textit{Overcooked}, we set $B = \{c, s, d, n, p\}$. To guarantee that we do not pass through a location with an object of type $B$, we will restrict any flow from leaving vertices assigned an object type in $B$.  By restricting flow from leaving blocking objects instead of entering them, we allow for flow to reach key objects $T$ that are also blocking objects $B$.

To complete our flow modeling, for each vertex $v \in V$ we create non-negative supply variables, $f^s_v \in \mathbb{Z}_{\geq0}$, and demand variables, $f^t_v \in \mathbb{Z}_{\geq0}$. These variables are enough to define a flow network between $S$ and $T$:

\begin{align}
    f^s_v &\leq \sum_{x \in S} |V| \cdot x_v \label{eq:connect_source}\\
    f^t_v &= \sum_{x \in T} x_v \label{eq:connect_target}\\
    f^s_v + \sum_{u:(u,v) \in E} f(u,v) &= 
    f^t_v + \sum_{u:(v:u) \in E} f(u,v) \label{eq:flow_constraint}\\
    f(u,v) + \sum_{x \in B}|V| \cdot x_u &\leq |V|,\ 
    \forall u:(u, v) \in E
    \label{eq:block_flow}
\end{align}

Equation \eqref{eq:connect_source} ensures that there is supply flow only in vertices in the space graph where a location $v \in V$ is assigned an object type from $S$. Note that multiple units of flow can leave source locations as we wish to reach many object types, but no more than $|V|$ locations exist with objects. Equation \eqref{eq:connect_target} creates exactly one unit of demand if location $v\in V$ is assigned an object from $T$. Equation \eqref{eq:flow_constraint} is a flow conservation constraint and ensures that flow entering a location $v \in V$ equals the flow leaving $v$. Equation \eqref{eq:block_flow} ensures that no flow leaves a location $v \in V$ that is assigned a blocking object.


In addition to reachability constraints, we introduce domain-specific constraints on the frequencies of objects of each type and ensure that neither agent can step outside the environment. First, we require that all locations on the border of the environment must be a blockable object type from $B$, and the environment contains exactly one robot $r$ and human $h$. Next, each environment requires at least one instance of a serve point $s$, an onion dispenser $n$, a dish dispenser $d$, and a pot $p$ to make it possible to fulfill food orders. Finally, we upperbound the number of serve points $s$, onion dispensers $n$, \mbox{dish dispensers $d$}, and \mbox{pots $p$} to reduce the complexity of the environment regarding runtime planning:

\begin{equation}
\begin{aligned}
    1 \leq \sum_{v \in V}s_v \leq 2 \qquad
    1 \leq \sum_{v \in V}n_v \leq 2 \\
    1 \leq \sum_{v \in V}d_v \leq 2 \qquad
    1 \leq \sum_{v \in V}p_v \leq 2 \\
\end{aligned}
\end{equation}
\begin{align}
    \sum_{v \in V}s_v + \sum_{v \in V}n_v + 
    \sum_{v \in V}d_v + \sum_{v \in V}p_v \leq 6
\end{align}

\noindent\textbf{Objective.} The constraints of our MIP formulation make any valid matching a solvable environment. To make the MIP a \textit{repair} method, we introduce a minimum edit distance objective from a provided input environment $K_i$ (in \textit{Overcooked} a kitchen)  to the repaired environment $K_r$ our MIP generates. 

We define an \textit{edit} as moving an object through a path in the space graph or changing which object type occupies a location. We wish to consider moving objects first, before changing the object type. Therefore, we permit different costs for different edit operations, with moving having a smaller cost than changing object type. A \textit{minimum cost edit repair} discovers a new environment $K_r$ that (1) minimizes the sum of costs for all edits made to convert $K_i$ to $K_r$, and that (2) satisfies all solvability constraints.

Following previous work~\cite{zhang2020video}, we formalize minimum cost repair as a minimum cost matching problem. Intuitively, we construct a matching between objects at locations in environment $K_i$ and objects at locations in environment $K_r$. Instead of considering all pairs of object locations between $K_i$ and $K_r$, we construct a matching as paths through the space graph $G$ from all objects in $K_r$ to all objects in $K_i$. Constructing matchings as paths allows us to assign costs based on the length of the path and corresponds to moving each object. 



Formally, we model our minimum cost matching as a minimum cost network flow problem\footnote{Note that these are separate networks than the one defined for the reachability problem.} for each object type. Consider creating a flow network for object type $o \in O$. First, we create a supply indicator $c_v$ (a constant). We assign $c_v = 1$ if and only if $K_i$ has an object of type $o$ to location $v$ and $c_v = 0$ otherwise. We then create a demand variable $f^t_v \in \{0, 1\}$ for each vertex $v\in V$. 



However, note there may not be a bijection between objects in $K_i$ and objects in $K_r$. To account for missing objects, we create \textit{waste variables} $r_v^t$, which consume flow not assigned to vertices with object type $o$. The separate variable enables the assignment of different costs to deletion and movement edits.


\begin{align}
    f^t_v &\leq o_v \label{eq:limit_demands}\\
    c_v + \sum_{u:(u,v)\in E} f(u,v) &= r^t_v + f^t_v + \sum_{u:(v,u)\in E} f(u,v) \label{eq:flow_constraint2}\\
    \sum_{v\in V}c_v &= \sum_{v\in V}f^t_v + \sum_{v\in V}r^t_v
    \label{eq:supply_demand}
\end{align}

Equation \eqref{eq:limit_demands} guarantees only vertices assigned objects of type $o$ have demands. Equation \eqref{eq:flow_constraint2} ensures flow conservation for each vertex and equation \eqref{eq:supply_demand} ensures that supplies and demands are equal.

The edit distance objective becomes minimizing the cost of deleting ($C_d = 20$) and moving ($C_m = 1$) objects:
\begin{equation}
    \sum_{o\in O} \Bigg( \sum_{v \in V}C_d r^t_v + \sum_{u,v:(u,v)\in E}C_m f(u,v) \Bigg)
\end{equation}

\noindent\textbf{MIP Implementation} We implement our MIP interfacing with IBM's CPLEX library~\citep{ibm_cplex}. Each MIP consists of 8850 variables and 3570 constraints to repair a $15 \times 10$ \textit{Overcooked} environment.

\section{QMDP Implementation} \label{sec:qmdp_implementation}

\begin{figure}[h]
    \centering
    \includegraphics[width=\linewidth]{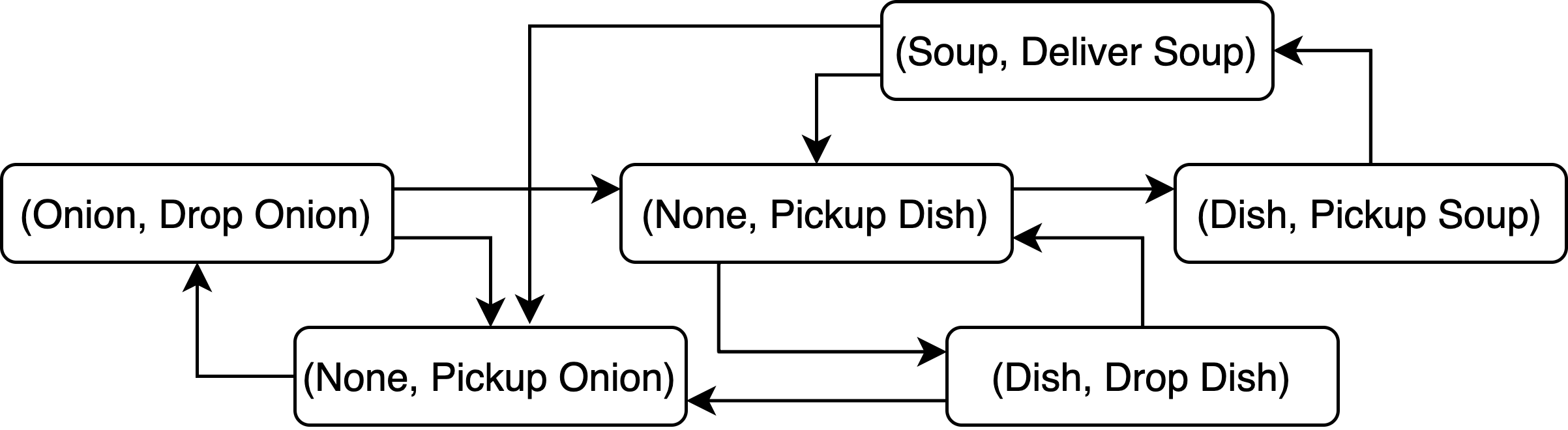}
    \caption{Human subtask state machine. The first element in the tuple is the object held by the simulated human; the second element is the subtask the human aims to complete.}
    \label{fig:human_state_machine}
\end{figure}

We specify the QMDP as a tuple $\{S, A, T, \Omega, O, C\}$:

\begin{itemize}
    \item $S$ is the set of states consisting of the observable and partially observable variables. The observable variables include the robot and human's held object, the number of items in a pot, and the remaining orders. The non-observable variable is the human subtask.
    \item $A$ is a set of robot subtasks that include picking up or dropping onion, dishes and soups. 
    \item $T: S \times A \rightarrow \Pi(S)$ is the transition function. We assume that the human does not change their desired subtask until it is completed and the environment is deterministic. Once a subtask is completed, the human chooses the next feasible subtask with uniform probability. 
    \item $\Omega$ is a set of observations. An observation includes the state of the world, e.g., number of onions in the pot, the human position and the current low-level human action (move up, down, left, right, stay, interact).
    \item $O:S \rightarrow \Pi(\Omega)$ is the observation function. Given a human subtask, the probability of an observation is proportional to the change caused by the human action in the cost of the motion plan to that subtask. This  makes subtasks more likely when the human moves towards their location. 
    \item $C: S \times A \rightarrow \mathbb{R}$ is the immediate cost. It is determined as the cost of the jointly optimal motion plan for the human and robot to reach their respective subtasks from their current positions.
\end{itemize}

Every time the human completes a subtask, the robot initializes its belief over all feasible subtasks with a uniform distribution, and updates its belief by observing the world after each human action. The feasible subtasks are determined by modeling the human subtask evolution with a state machine, shown in Fig.~\ref{fig:human_state_machine}.

The best action is selected by choosing the robot subtask with the highest expected value $Q(s,a) = \mathbb{E}[C(s,a)+V(s')]$, given the current belief on the human subtask.


\begin{algorithm}[t]
 \caption{Latent Space Illumination with CMA-ME}
\label{algorithm: cmame}
\begin{algorithmic}
\STATE\textbf{Input:} GAN Generator Network $G$
\STATE\textbf{Initialize:} archive of latent vectors $\mathcal{X}\leftarrow \emptyset$ and assessments $\mathcal{F}\leftarrow \emptyset$

\STATE Initialize population of emitters $E$ with initial Covariances $C_e$ and means $\bm{\mu_e}$.
\FOR{$t=1,\ldots,N $}
\STATE Select emitter $e$ from $E$
\STATE Generate vector $\bm{z}  \leftarrow \mathrm{generate(\bm{\mu_e}, C_e)}$ 
\STATE Environment $Env \leftarrow G(\bm{z})$
\STATE Environment $Env' \leftarrow \mathrm{MIP
\_repair} (Env)$
\STATE Simulate agents in $Env'$
\STATE $f \leftarrow \mathrm{performance}(Env')$
\STATE $\bm{b}  \leftarrow \mathrm{behaviors}(Env')$
\STATE $\mathrm{emitter\_update}(e, \bm{z}, f, \bm{b})$
 \IF{$\mathcal{F}[\bm{b}] = \emptyset$ or $\mathcal{F}[\bm{b}] < f$ } 
 \STATE Update archive $\mathcal{X}[\bm{b}] \leftarrow \bm{z},\mathcal{F}[\bm{b}] \leftarrow f$   
 \ENDIF
\ENDFOR
\end{algorithmic}
\label{alg:map-elites}
\end{algorithm}

\begin{algorithm}[t]
 \caption{emitter\_update (Improvement emitter)}
\label{algorithm: improvement emitter}
\begin{algorithmic}
\STATE\textbf{Input:} emitter $e$, latent vector $\bm{z}$, performance $f$, and behaviors $\bm{b}$
\STATE Unpack the parents, sampling mean $\bm{\mu_e}$, covariance matrix $C_e$, and parameter set $P_e$ from $e$.
\IF{$\mathcal{X}[\bm{b}]=\emptyset$}
\STATE $\Delta \leftarrow f$
\STATE Flag that $\bm{z}$ discovered a new cell
\STATE Add $\bm{z}$ to parents 
\ELSE
\IF{$\mathcal{F}[\bm{b}]<f$}
\STATE $\Delta \leftarrow f - \mathcal{F}[\bm{b}]$
\STATE Add $\bm{z}$ to parents 
\ENDIF 
\ENDIF
\IF{sampled population is size $\lambda$}
\IF{parents $\neq \emptyset$}
\STATE Sort parents by their $\Delta$ values (prioritizing parents that discovered new cells)
\STATE Update $\bm{\mu_e}, C_e, P_e$ by parents
\STATE $\textrm{parents} \leftarrow
\emptyset$
\ELSE
\STATE Restart from random elite in $\mathcal{X}$
\ENDIF
\ENDIF
\end{algorithmic}
\label{alg:map-elites}
\end{algorithm}
\section{Latent Space Illumination with CMA-ME} \label{sec:LSI-CMAME}

Algorithm~\ref{algorithm: cmame} presents the \mbox{CMA-ME} algorithm, adapted for latent space illumination. \mbox{CMA-ME} consists of three components: a population of emitters, a scheduling algorithm for emitters, and an archive. Each emitter is a modified \mbox{CMA-ES} instance focused on improving a different area of behavior space. The scheduling algorithm selects emitters in a round-robin fashion to generate new environments.

To generate an environment from an emitter, we sample a latent vector for the emitter's Gaussian distribution $\mathcal{N}(\bm{\mu_e}, C_e)$. We then pass that latent vector to the GAN generator network to generate a new candidate environment, after which we repair the environment with the MIP solver. Next, we simulate the human and robot in the environment and compute performance $f$ and behavior characteristics $\bm{b}$. We compute which unique cell the new environment belongs to based on $\bm{b}$, and we insert the new environment into the archive if the new solution improved upon the current environment occupying the cell with respect to $f$. Information about how the archive changes when we insert the new environment governs how we update an emitter's mean $\bm{\mu_e}$ and covariance $C_e$. Conceptually, $\mathcal{N}(\bm{\mu_e}, C_e)$ models which directions from position $\bm{\mu_e}$ in latent space result in the largest changes in the archive.


The authors of \mbox{CMA-ME} proposed several types of emitters. In this paper, we configure \mbox{CMA-ME} with \textit{improvement emitters} since they were shown to offer good balance between quality and diversity. Algorithm~\ref{algorithm: improvement emitter} details how \mbox{CMA-ME} updates improvement emitters based on how the archive changes. After generating and evaluating a new environment, we return the environment to the emitter with performance $f$ and behavior characteristics $\bm{b}$. When adding environments to the archive, there are two cases which change the archive: the environment occupies a new cell in the archive or the environment improves upon an existing cell. For new cells, the archive change $\Delta$ is the raw performance $f$. If the environment improves upon an existing cell, the archive change $\Delta$ is the difference between performance $f$ and the performance of the incumbent environment $\mathcal{F}[\bm{b}]$. For each new candidate environment that changes the archive we store that environment in a list called \textit{parents}. Once the number of sampled environments reaches $\lambda$ samples, we have enough samples to update $\mathcal{N}(\bm{\mu_e}, C_e)$. First, we sort parents by two criteria: newly discovered cells have higher priority than improvements to existing cells and next by the computed $\Delta$ values. The ranking corresponds to a log-likelihood estimate on which search directions are most profitable (see~\cite{brookes2020view}), and we adapt the emitter's mean $\bm{\mu_e}$, covariance $C_e$, and additional CMA-ES parameters $P_e$\footnote{$P_e$ contains evolution paths and the global step-size scalar $\sigma$. For more information on CMA-ES's adaptation mechanisms see \cite{hansen:cma16}.} towards search directions which yielded the largest archive improvements. If the archive didn't change after $\lambda$ samples, then the emitter restarts from a latent code in the archive chosen uniformly at random as $\bm{\mu_e}$ and an identity covariance matrix $C_e = I$.

\section{Implementation of Latent Space Illumination}  \label{sec:implementation}

\subsection{GAN Training} \label{subsec:gan-training}

We hand-authored \textit{Overcooked AI} environments based on $77$ levels from the \textit{Overcooked}~\citep{overcooked} and \textit{Overcooked 2}~\citep{overcooked2} video games.\footnote{We include the hand-authored environments as supplemental material.} For training the GAN, we augment the environments by mirroring each environment horizontally, vertically, and both horizontally and vertically to expand the training set to \num{308} training environments in total. 


We encode each environment as a \num{2}-dimensional grid of tokens, where each token represents different object types (e.g., countertop, pot and onion dispenser). We then convert each environment layout to \num{2}-dimensional tensor where each cell contains an \num{8}-dimensional vector with the one-hot encoding of the tile. We then pad the layout to a $8 \times 16 \times 16$ input tensor which gets passed to the discriminator network. For the generator network, we make the latent vector input \num{32}-dimensional, following previous work~\cite{fontaine2020illuminating, volz2018evolving}.


We train the GAN with RMSprop for \num{50000} iterations with learning rate of $1e^{-5}$ and a batch size of \num{64} in approximately \num{2} hours.

\subsection{CMA-ME} \label{subsec:cmame}
A single run of CMA-ME deploys \num{5} improvement emitters with population $\lambda = 37$, and mutation power $\sigma = 0.2$. We ran all experiments for \num{10000} evaluations in a distributed manner on a high performance cluster, utilizing \num{120} Xeon CPUs. Each experiment took between \num{12} and \num{18} hours to complete.

\subsection{Performance function} \label{subsec:performance}
We wish to represent in the performance function $f$ both the number of completed orders and the amount of time each order was completed. Thus, we use a rank-based aggregate function with at most five digits. The fifth digit (in the ten-thousand place) indicates the number orders delivered. The fourth and the third digit indicate the number of time steps left after the second order is delivered (the simulation ended after 100 timesteps), and the second digit and first digit indicate the number of time steps left after the first order is delivered. For example, if \num{2} orders are delivered with the first one at time step \num{20} and the second \num{60}, the performance value would be \num{24080}. If no order is delivered, the performance is \num{0}. Finally, we normalize the value so that it is between \num{0} and \num{1}.

\subsection{Multiple Trials} \label{subsec:trials}
When running multiple trials (section~\ref{subsec:human-aware}), we compute the median performance to estimate the performance metric $f$, since the  performance metric is rank-based. Similarly, we used the median, rather than the mean, for the estimation of BCs, since the median is more robust to outliers, and we wanted to avoid the case that an environment where one agent does all of the work in nearly half of the trials and none of the work in the other half would be placed in the ``even workload'' region of the archive. 

We ran \num{50} trials for the median calculation, since we empirically found this number of trials reduced variance enough to give consistent results.



\section{Alternatives to Latent Space Illumination} \label{subsec:alternatives}
\subsection{Random Search} \label{sec:random-search}
Previous work~\cite{fontaine2020illuminating} demonstrates that searching the latent space of a GAN with CMA-ME significantly outperforms random search in the quality and diversity of generated levels for the \emph{Super Mario Bros.} video game. In random search, we directly sample the latent space of the GAN from the same distribution that we used to train the generator network: a normal distribution with zero mean and variance equal to one. For each sample, we generate an environment, repair the environment with the MIP solver and update the archive, according to the \mbox{CMA-ME} algorithm.

\begin{figure}[t!]
\centering
    \begin{subfigure}{1.0\columnwidth}
    \includegraphics[width=\linewidth]{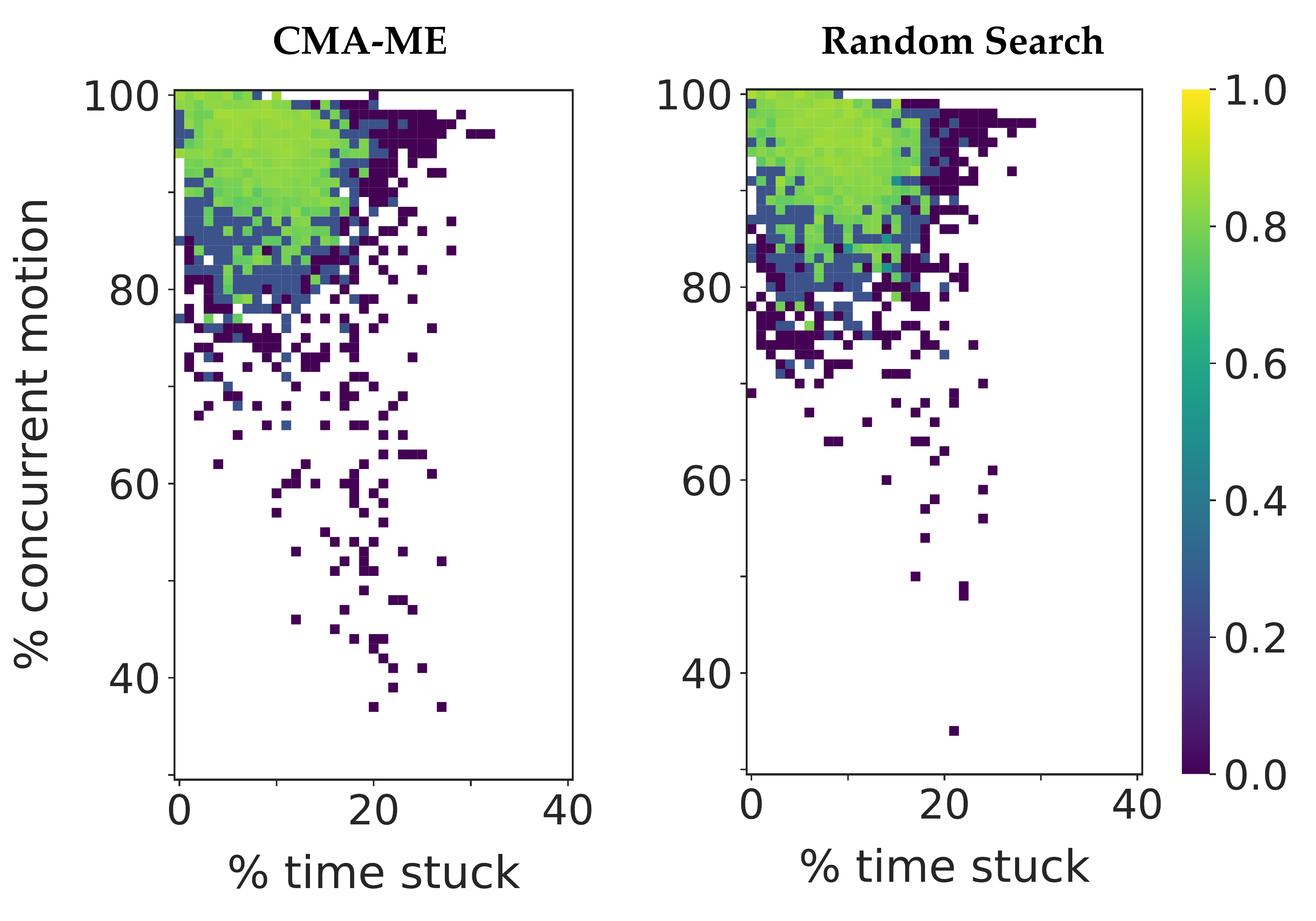}
    \caption{Archives generated after \num{10000} evaluations.}
    \label{fig:CMAMEvsRandom_10k}
    \end{subfigure}
    \begin{subfigure}{1.0\columnwidth}
    \includegraphics[width=\linewidth]{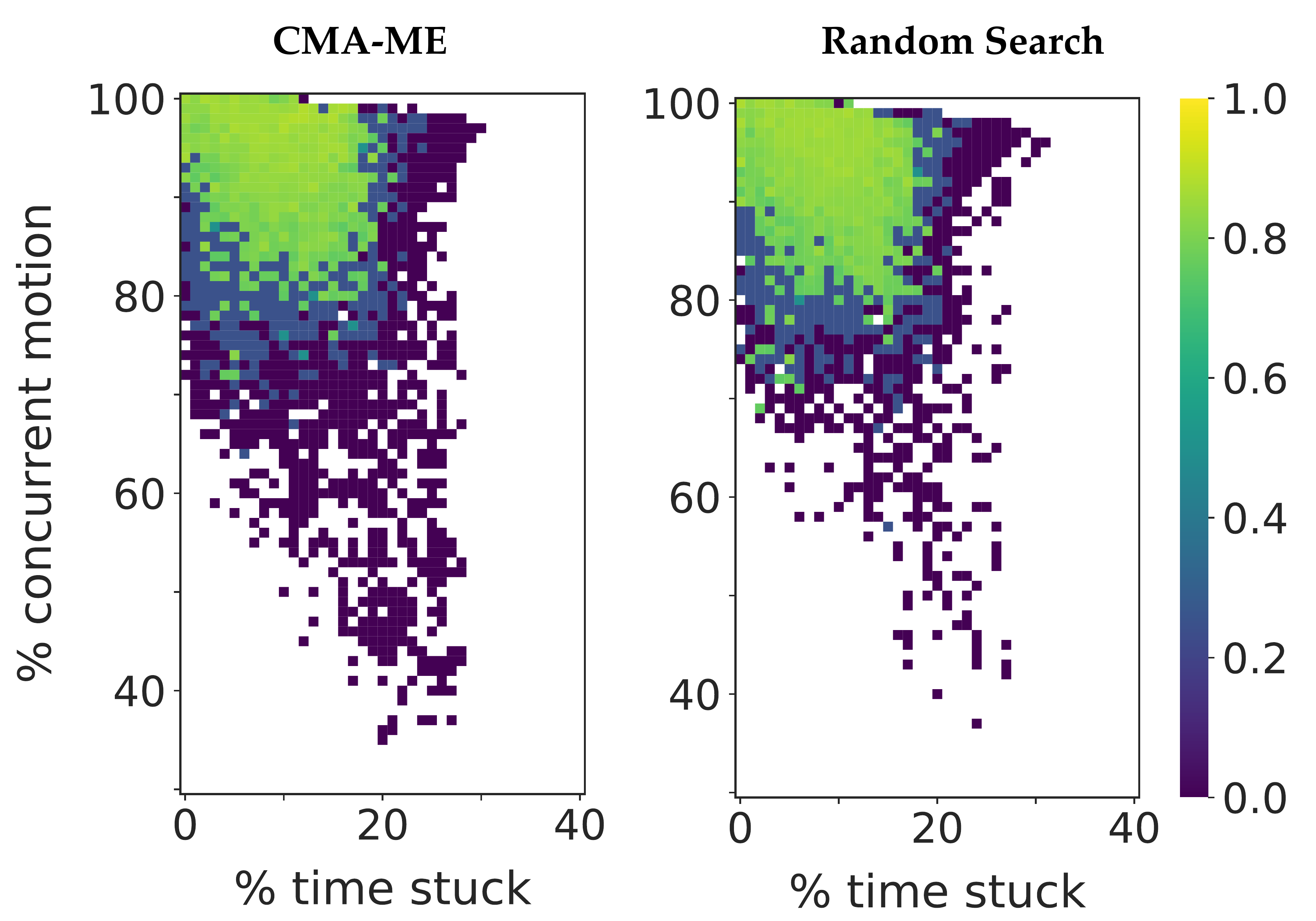}
    \caption{Archives generated after \num{50000} evaluations.}
            \label{fig:CMAMEvsRandom_50k}
    \end{subfigure}
\caption{Comparison of the team-fluency archives generated with CMA-ME and random search.}
\label{fig:CMAMEvsRandom}
\end{figure}

As a proof of concept, we show the generated archive for the team fluency BCs for a single run of random search. After \num{10000} evaluations, random search populates the archive with \num{563} unique environments, compared to \num{624} by CMA-ME. We observe that, while the difference between the two methods is smaller compared to previous experiments in the \emph{Super Mario Bros.} domain~\cite{fontaine2020illuminating}, CMA-ME visibly finds more environments in the bottom half of the archive (Fig.~\ref{fig:CMAMEvsRandom_10k}). These environments have unique characteristics such as narrow corridors leading to dead-ends and are hard to find by directly sampling GAN latent vectors from a fixed distribution. This result shows the capability of CMA-ME to find environments at extreme positions of the behavior space.

Since both methods are stochastic, we compare CMA-ME and random search by doing \num{5} runs of CMA-ME and \num{5} runs of random search. A one-way ANOVA found the difference in coverage to be significantly larger for CMA-ME ($F(1,8) = 10.064, p = 0.013)$, supporting previous findings~\cite{fontaine2020illuminating}.

Fig.~\ref{fig:CMAMEvsRandom_50k} shows the archives of the two algorithms after \num{50000} evaluations. The difference between the two methods is more visible, since CMA-ME has more time to adapt towards the hard-to-reach regions of the behavior space.

\begin{figure}
\centering
\begin{tabular}{cc}
\begin{subfigure}{0.2\textwidth}
    \includegraphics[width=\linewidth]{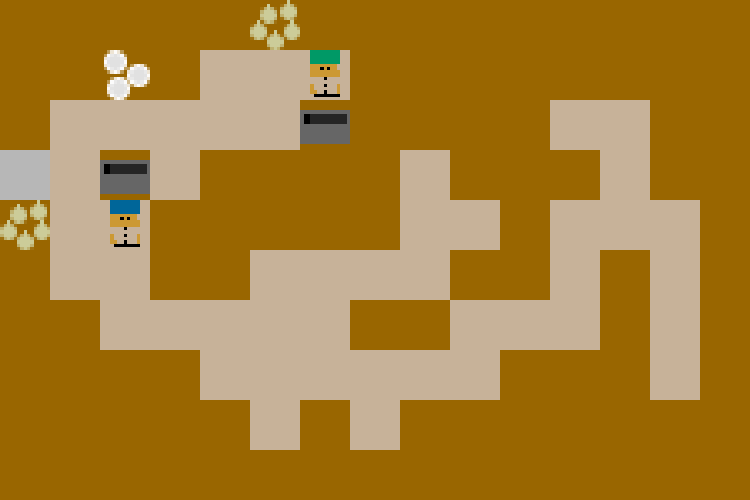}
    \caption{}
    \label{fig:my_label}
\end{subfigure}&
\begin{subfigure}{0.2\textwidth}
    \includegraphics[width=\linewidth]{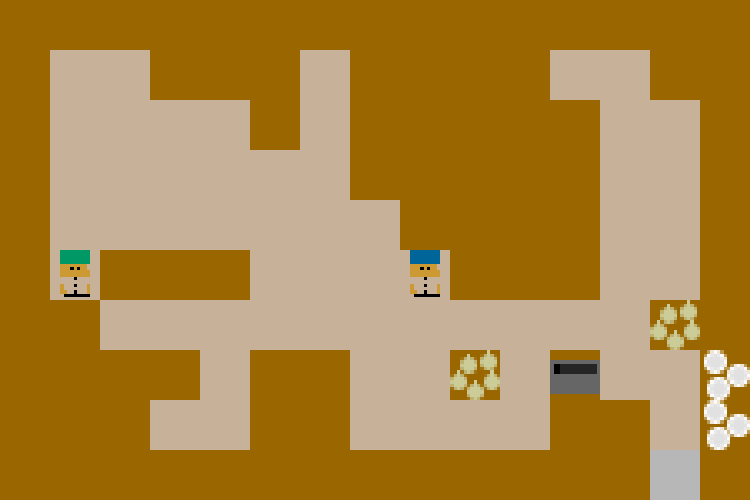}
    \caption{}
    \label{fig:my_label}
\end{subfigure}\\
\end{tabular}
\caption{Example environments generated with a MAP-Elites implementation that directly searches over the tiles of the layouts, rather than searching the latent space of a GAN.}
\label{fig:baseline}
\end{figure}

\subsection{Directly Searching for Environments}
\label{sec:map_elites_baseline}
To demonstrate the benefit of searching the latent space of a GAN, we implemented a baseline vanilla MAP-Elites (ME) algorithm~\cite{mouret2015illuminating}, where we directly search over layout tiles. We choose MAP-Elites, instead of CMA-ME, since CMA-ME operates only on continuous search spaces. We ran ME for \num{10000} iterations. At each iteration an environment was selected randomly from the archive. For each selected environment, we generated a new environment by randomly selecting \num{20} tiles and replacing them with new ones, the type of which were sampled uniformly. We then repaired and evaluated each new environment using the workload distribution BCs and the centralized planners, identically to the GAN-generated environments.

Fig.~\ref{fig:baseline} shows two example environments, one of even and one of uneven workload, generated by this method. We observe that the two environments are stylistically different from the human-authored examples: they include empty rooms and they lack the partial symmetry often exhibited in the human-authored examples and the GAN-generated environments.

\section{User Study} \label{sec:study2}

\begin{figure}[t!]
\centering
\begin{subfigure}{0.5\textwidth}
    \includegraphics[width=\linewidth]{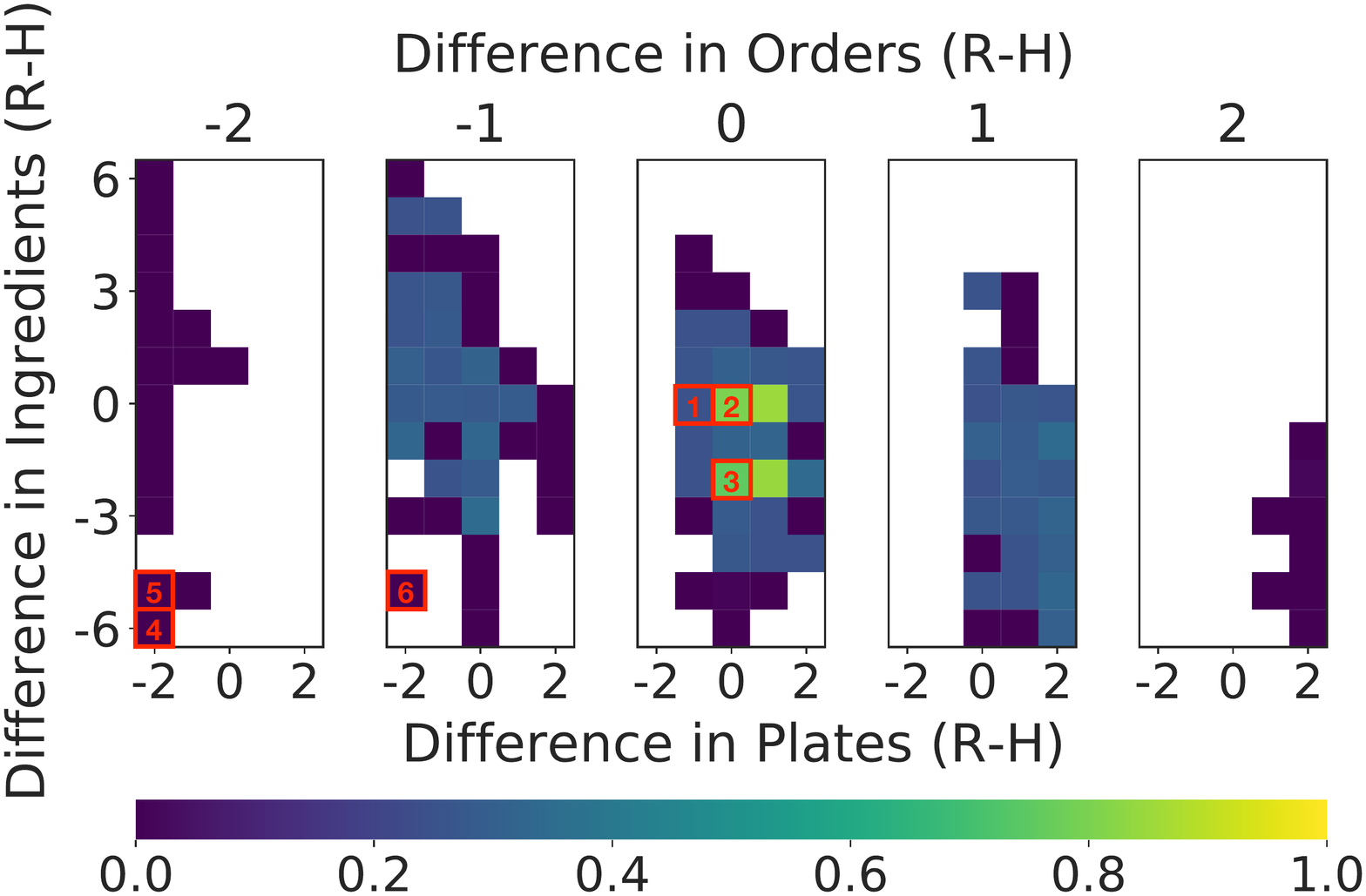}
\end{subfigure}\hfil
\begin{tabular}{cc}
\begin{subfigure}{0.2\textwidth}
    \includegraphics[width=\linewidth]{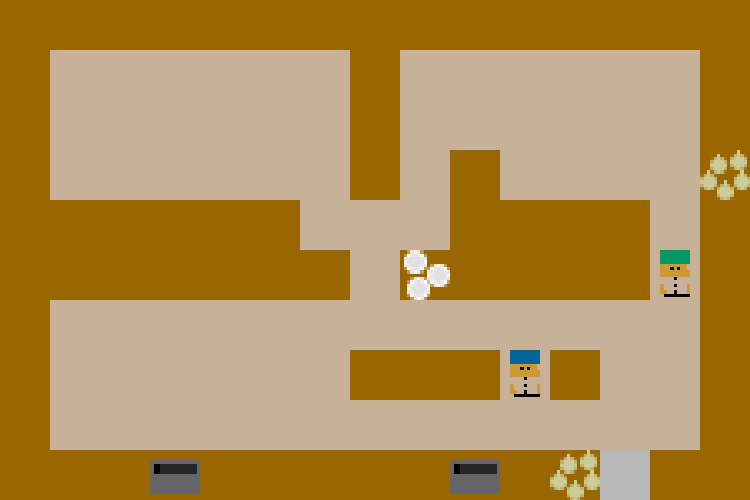}
    \caption*{(1)}
\end{subfigure}&
\begin{subfigure}{0.2\textwidth}
    \includegraphics[width=\linewidth]{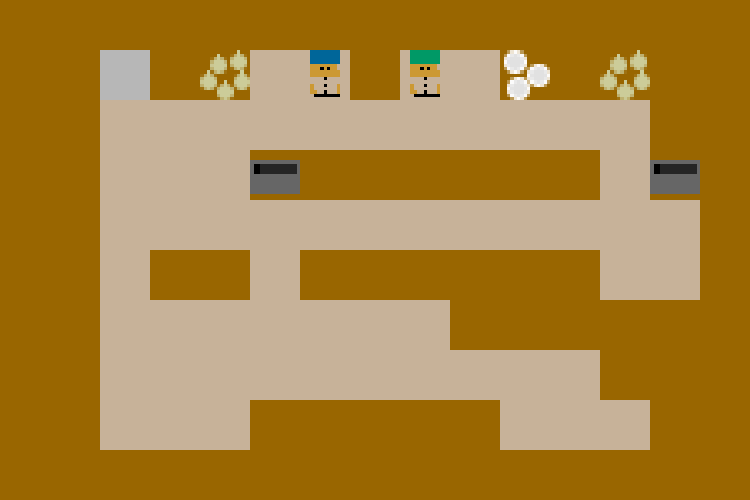}
    \caption*{(2)}
\end{subfigure}\\
\begin{subfigure}{0.2\textwidth}
    \includegraphics[width=\linewidth]{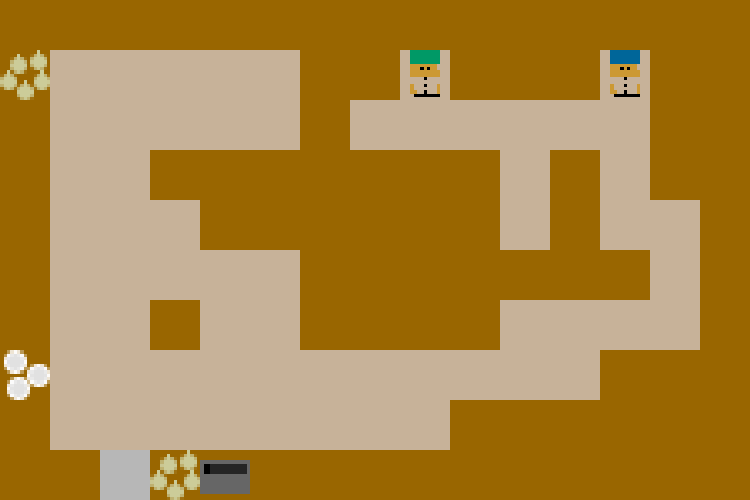}
    \caption*{(3)}
\end{subfigure}&
\begin{subfigure}{0.2\textwidth}
    \includegraphics[width=\linewidth]{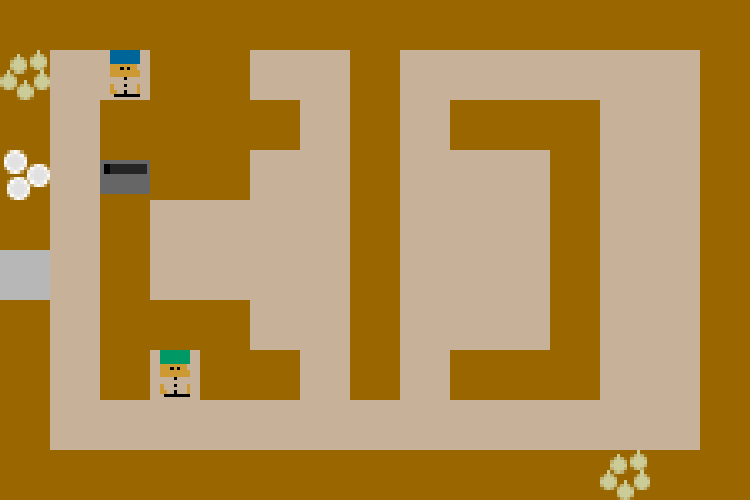}
    \caption*{(4)}
\end{subfigure}\\
\begin{subfigure}{0.2\textwidth}
    \includegraphics[width=\linewidth]{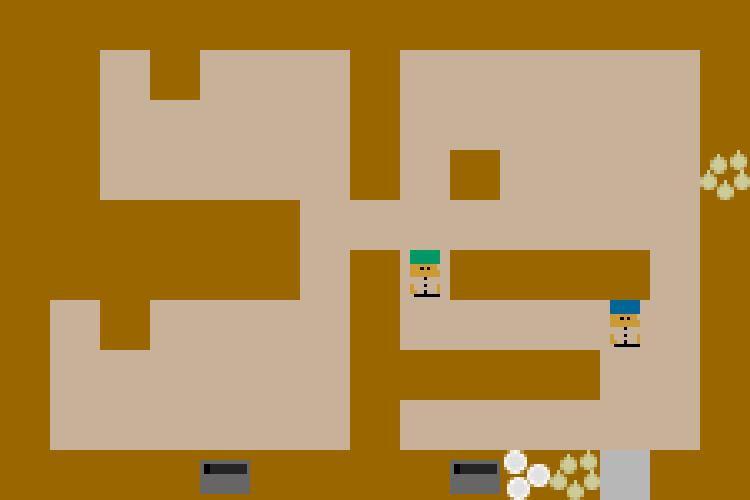}
    \caption*{(5)}
\end{subfigure}&
\begin{subfigure}{0.2\textwidth}
    \includegraphics[width=\linewidth]{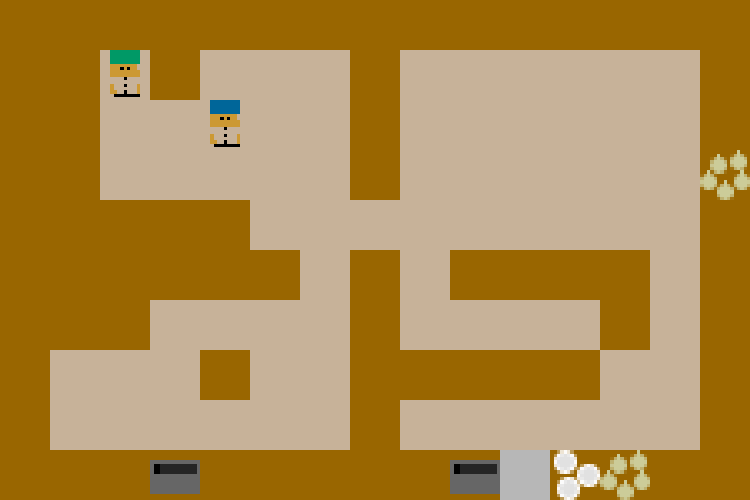}
    \caption*{(6)}
\end{subfigure}\\
\end{tabular}
\caption{Environments selected for the user study. (1)-(3) are ``even workload'', (4)-(6) are ``uneven workload.''}
\label{fig:user_study_workload}
\end{figure}

\begin{figure}[t!]
\centering
\begin{subfigure}{0.33\textwidth}
    \includegraphics[width=\linewidth]{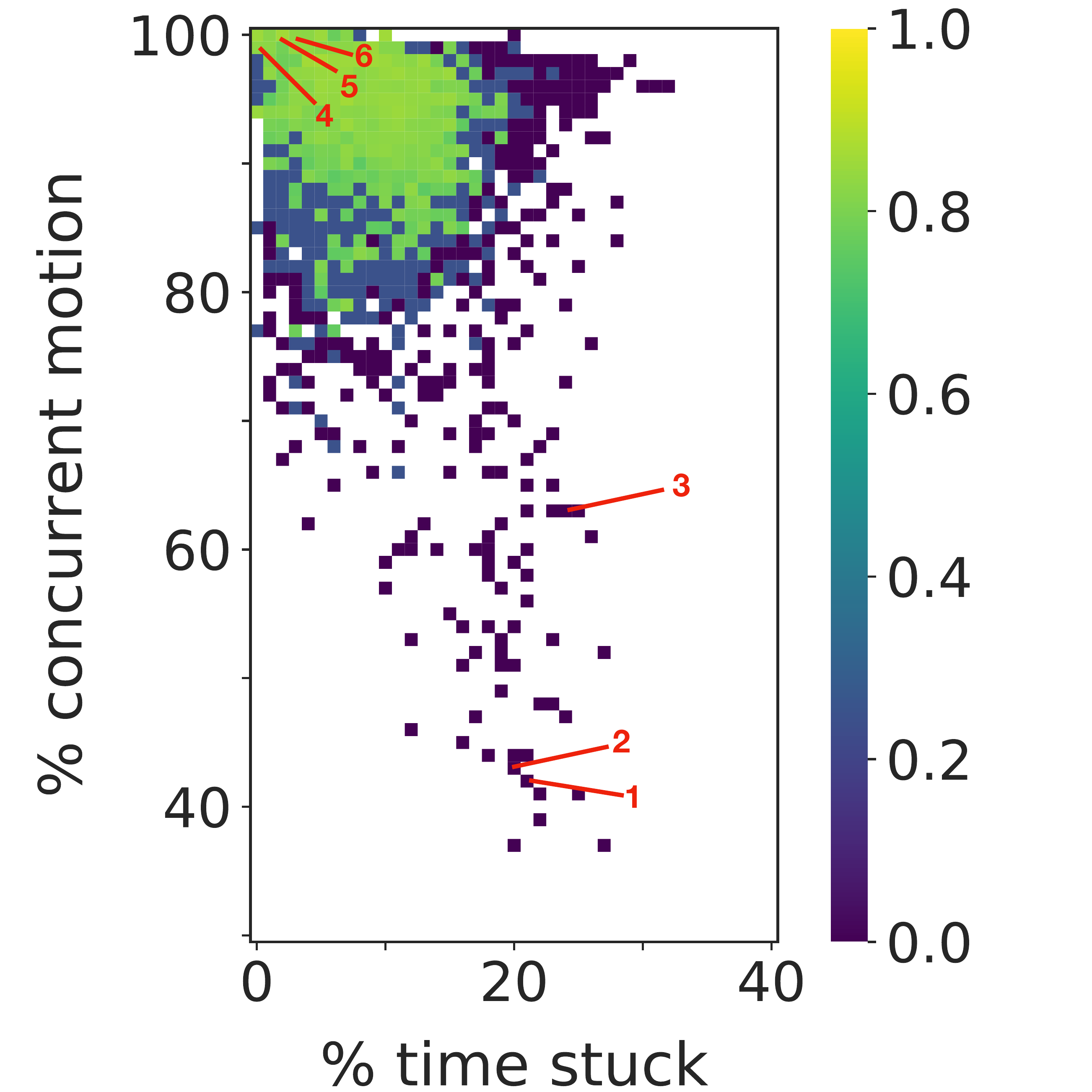}
\end{subfigure}\hfil
\vspace{1em}
\begin{tabular}{cc}
\begin{subfigure}{0.2\textwidth}
    \includegraphics[width=\linewidth]{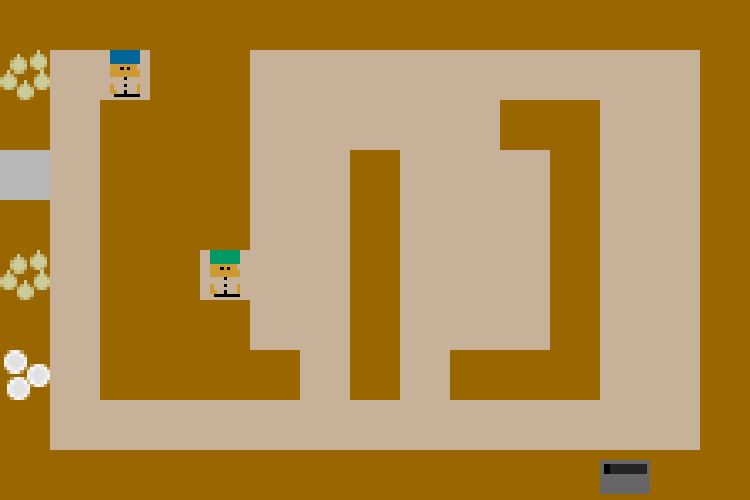}
    \caption*{(1)}
    \label{fig:qmdp_min_map_1}
\end{subfigure}&
\begin{subfigure}{0.2\textwidth}
    \includegraphics[width=\linewidth]{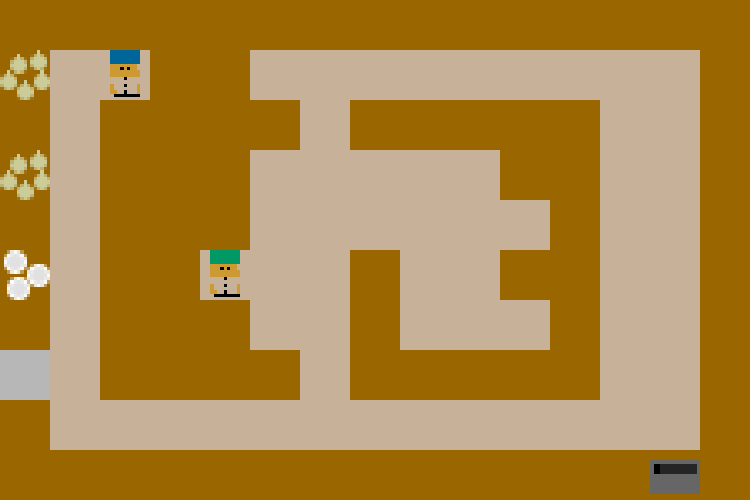}
    \caption*{(2)}
    \label{fig:qmdp_min_map_2}
\end{subfigure}\\
\begin{subfigure}{0.2\textwidth}
    \includegraphics[width=\linewidth]{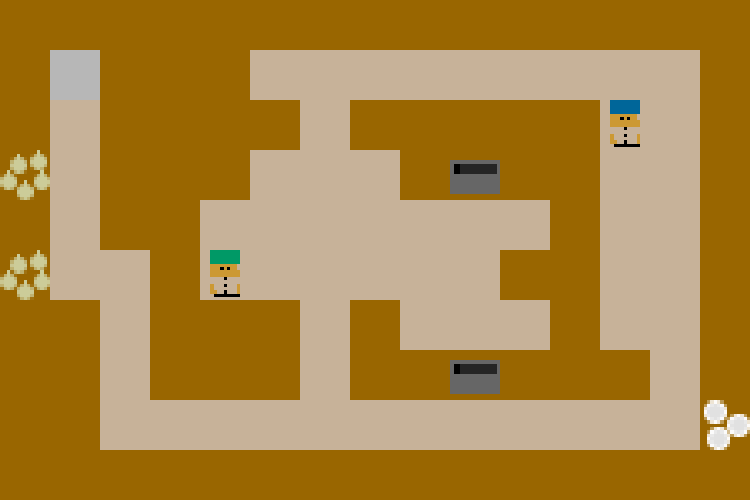}
    \caption*{(3)}
    \label{fig:qmdp_min_map_1}
\end{subfigure}&
\begin{subfigure}{0.2\textwidth}
    \includegraphics[width=\linewidth]{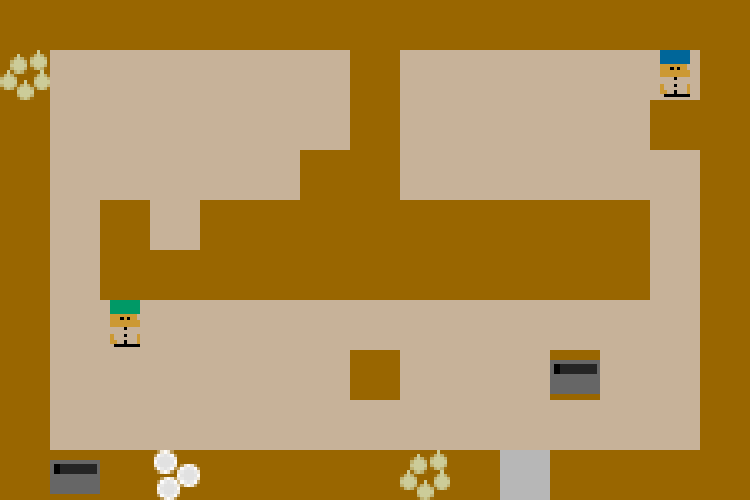}
    \caption*{(4)}
    \label{fig:qmdp_min_map_2}
\end{subfigure}\\
\begin{subfigure}{0.2\textwidth}
    \includegraphics[width=\linewidth]{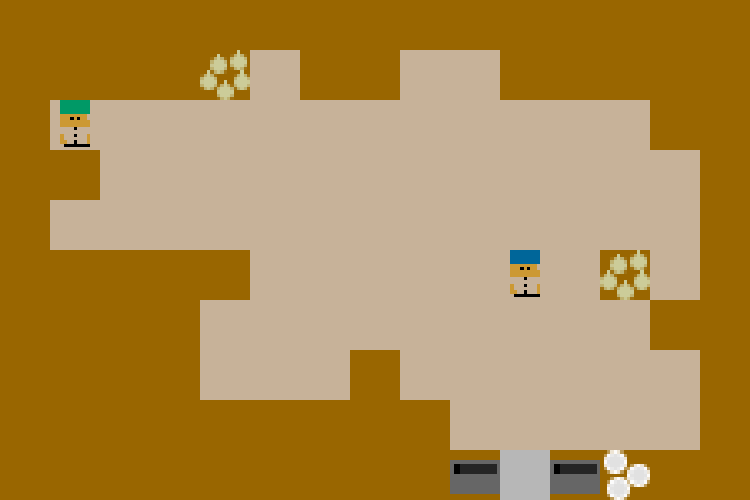}
    \caption*{(5)}
    \label{fig:qmdp_min_map_1}
\end{subfigure}&
\begin{subfigure}{0.2\textwidth}
    \includegraphics[width=\linewidth]{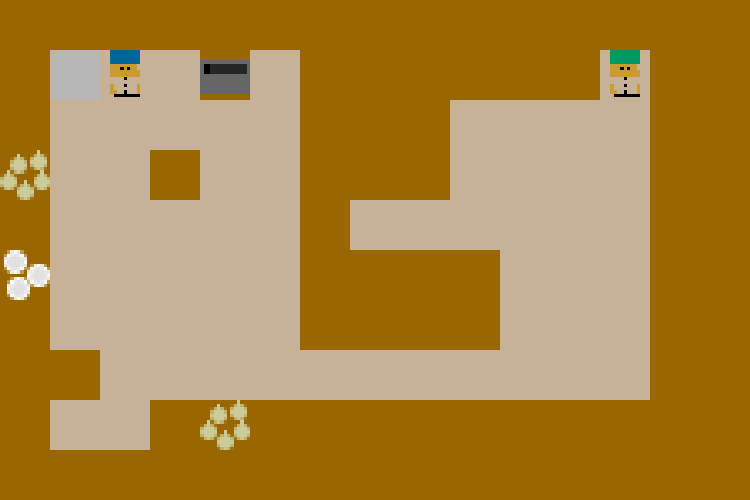}
    \caption*{(6)}
    \label{fig:qmdp_min_map_2}
\end{subfigure}\\
\end{tabular}
\caption{Environments selected for the user study. (1)-(3) are ``low team fluency'', (4)-(6) are ``high team fluency.''}
\label{fig:user_study_fluency}
\end{figure}

\begin{figure}[t!]
\centering
\begin{subfigure}{0.9\columnwidth}
    \includegraphics[width=\columnwidth]{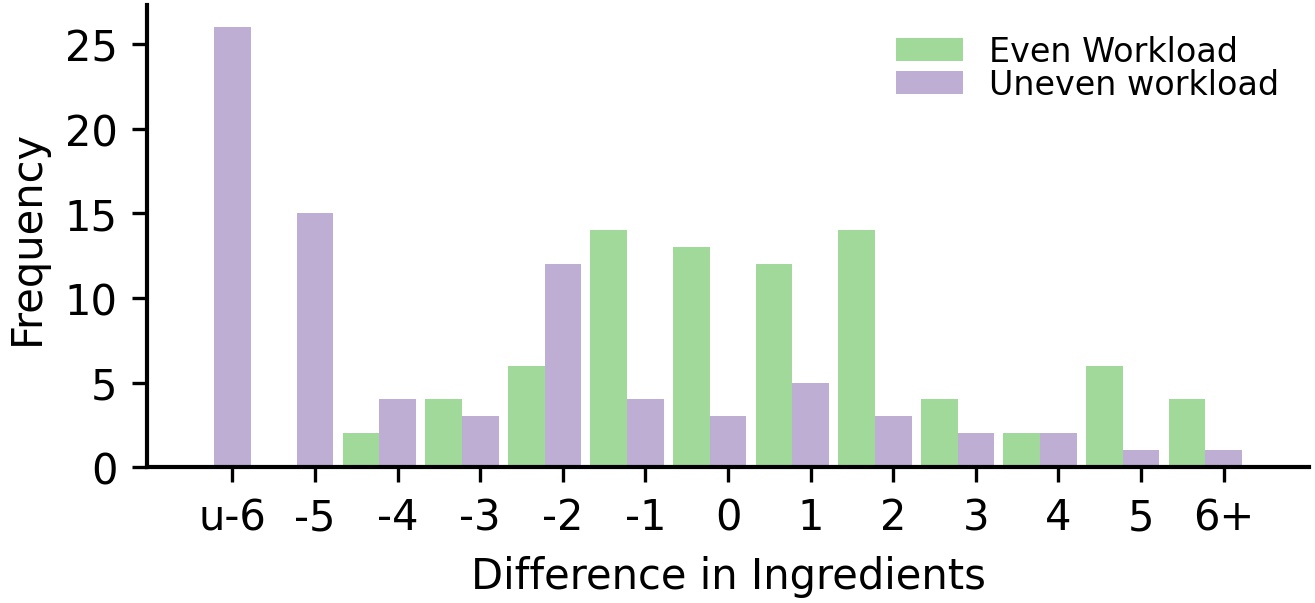}
\end{subfigure}
\begin{subfigure}{1.0\columnwidth}
    \includegraphics[width=\columnwidth]{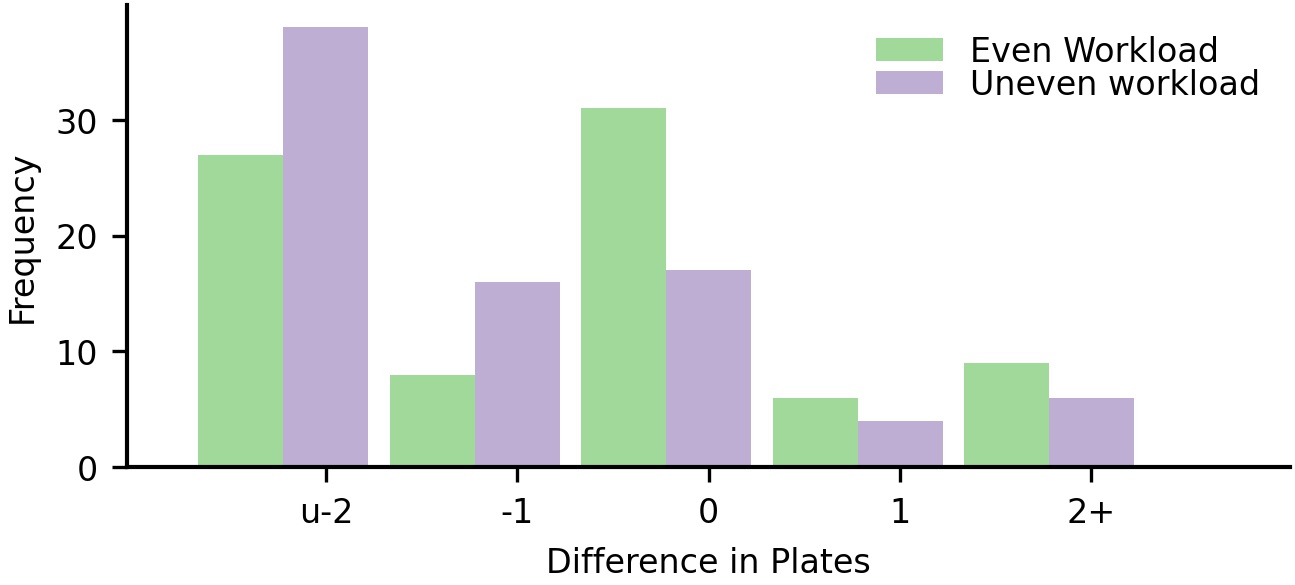}
\end{subfigure}
\begin{subfigure}{1.0\columnwidth}
    \includegraphics[width=\columnwidth]{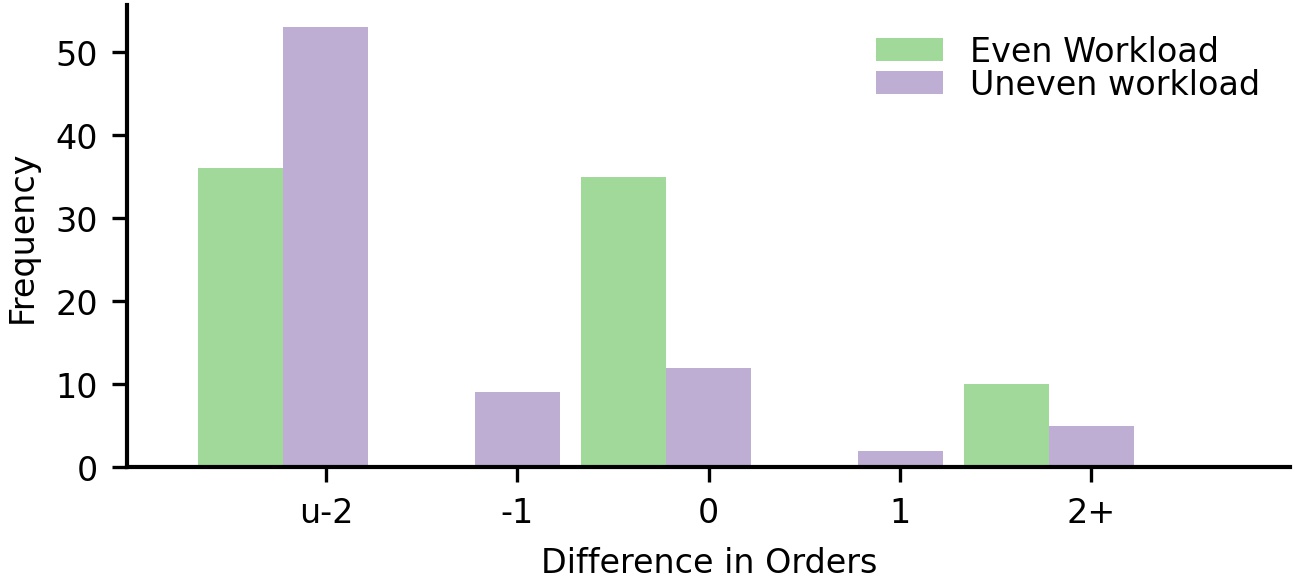}
\end{subfigure}
\caption{Histograms of computed BCs from the playthroughs of the users for even and uneven workload environments.}
\label{fig:hist_workload}
\end{figure}

\begin{figure}[t!]
    \centering
    \begin{subfigure}{1.0\columnwidth}
    \includegraphics[width=\columnwidth]{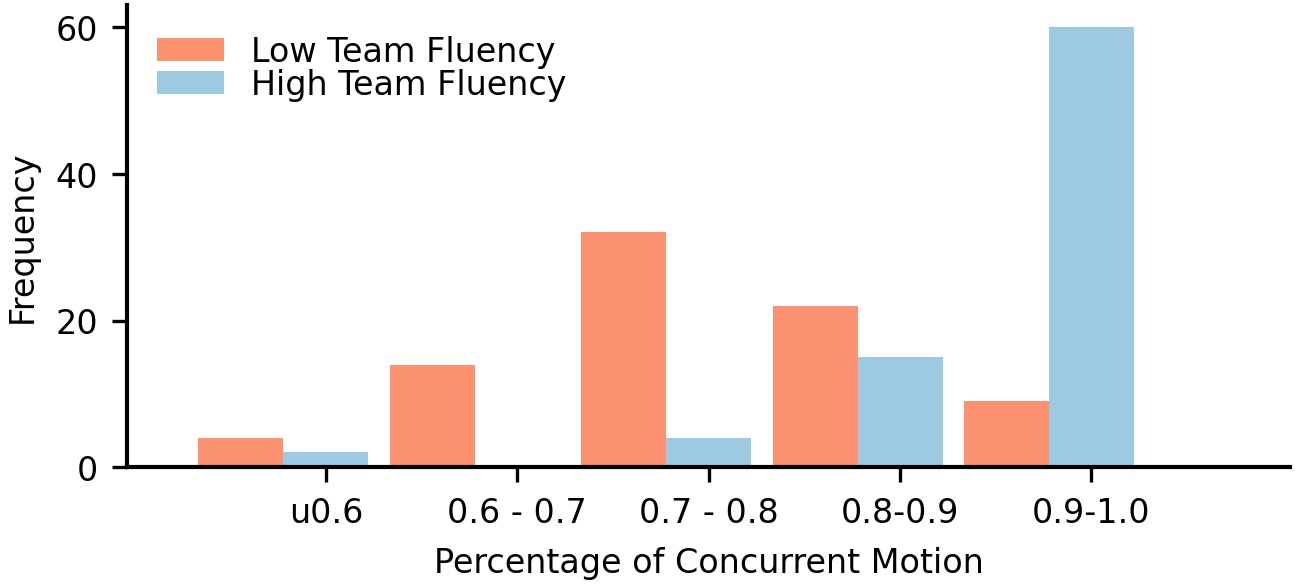}
    \end{subfigure}
    \begin{subfigure}{1.0\columnwidth}
    \includegraphics[width=\columnwidth]{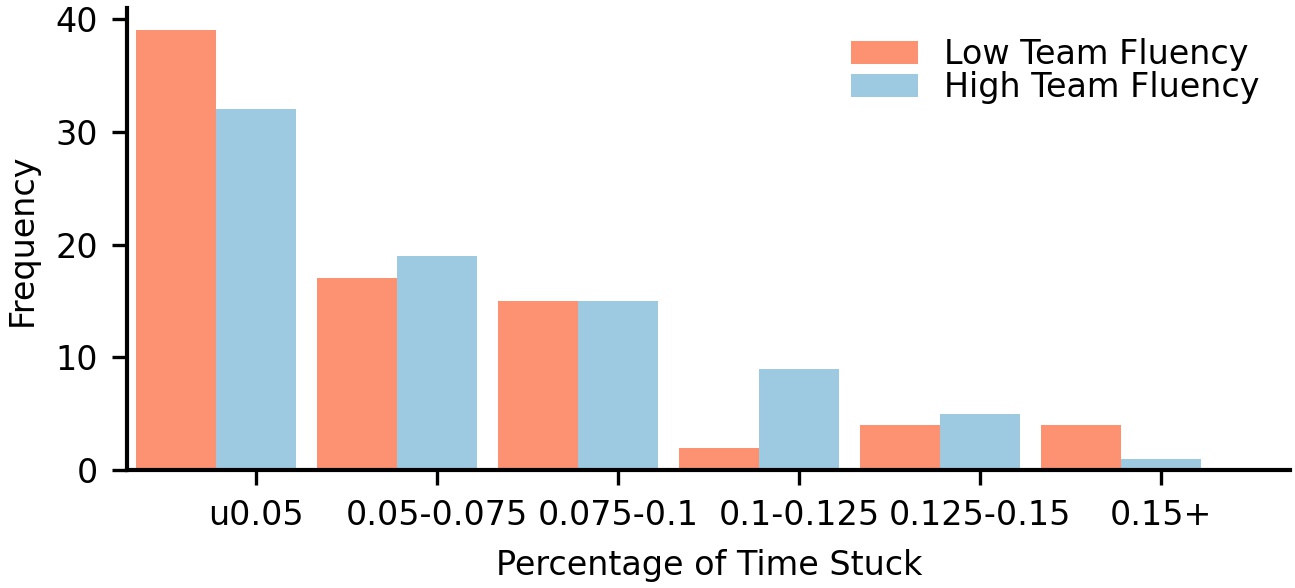}
    \end{subfigure}
    \caption{Histograms of computed BCs from the playthroughs of the users for the low and high team fluency environments.}
    \label{fig:hist_fluency}
\end{figure}

\subsection{Environments}
Fig.~\ref{fig:user_study_workload} and \ref{fig:user_study_fluency} show the environments selected for the user study. We selected environments from the archive of Fig.~\ref{fig:qmdp_min_map}, since these environments were more challenging for coordination.

\subsection{Histograms of Computed Behavior Characteristics}

Fig.~\ref{fig:hist_workload} and \ref{fig:hist_fluency} show the computed BCs from the playthroughs of the users for even and uneven workload environments, as well as for low and high team fluency environments.  

\subsection{Open-ended Responses}
In a post-experiment questionnaire, we asked participants to describe which factors affected the quality of the collaboration. In response, participants noted the ``openness of the space,'' the ``number of pots'' and the ``location of the stove, onions and delivery port.'' They also referred to the robot's behavior.  According to one participant, the robot recognized that ``I am walking in a narrow pathway and stand aside to not block my way. This really helps the collaboration because I can understand that the robot is trying to help me by standing still.'' Others stated that ``sometimes the robot blocked my way,''  ``Sometimes we did the same next step which makes the procedure less efficient'' and ``[the robot] does not seem to know how to `pass' items to me (take items and put them somewhere so I can pick them up).'' The latter is an ability that we excluded from the QMDP model to enable real-time computation of the QMDP policy. 

We additionally asked participants to describe which features made some environments more challenging than others. Most participants stated that long and narrow paths made environments harder, especially when there were no alternative routes. A few participants also pointed to the distance between objects as a factor, which affected the ability of the team to complete the task in time.

Overall, in addition to the robot's performance, participants recognized the environment as a significant factor influencing the quality of the collaboration. This supports the importance of searching for diverse environments when studying coordination behaviors. 




\end{document}